\ifdefined\SINGLECOLUMN
  \documentclass[11pt,onecolumn,journal,compsoc]{IEEEtran}
\else
  \documentclass[10pt,journal,compsoc]{IEEEtran}
\fi
\ifCLASSOPTIONcompsoc
  \usepackage[nocompress]{cite}
\else
  \usepackage{cite}
\fi
\ifCLASSINFOpdf
\else
\fi
\ifCLASSOPTIONcompsoc
  \usepackage[caption=false,font=footnotesize,labelfont=sf,textfont=sf]{subfig}
\else
  \usepackage[caption=false,font=footnotesize]{subfig}
\fi
\usepackage{multirow}
\usepackage[colorlinks=true,
            citecolor=blue,
            linkcolor=blue,
            urlcolor=blue]{hyperref}
\usepackage{graphicx}
\makeatletter
\newcommand{\refmain}[1]{%
  \ifcsname r@main#1\endcsname
    \ref{main#1}%
  \else
    \ref{#1}%
  \fi
}
\makeatother

\usepackage{orcidlink}
\usepackage{amsmath}
\usepackage{amssymb}
\usepackage{booktabs}
\usepackage{threeparttable}
\usepackage{etoolbox}
\usepackage{xcolor}
\definecolor{bestcolor}{HTML}{F5E56E}
\definecolor{secondcolor}{HTML}{A8D8F0}
\newcommand{\best}[1]{\colorbox{bestcolor}{\textbf{#1}}}
\newcommand{\second}[1]{\colorbox{secondcolor}{#1}}
\definecolor{revisionblue}{RGB}{0,70,180}

\ifdefined\TRACKEDCHANGES
  \newcommand{\revisionon}{\color{revisionblue}}
  
\else
  \newcommand{\revisionon}{}
  
\fi

\begin{document}
%\title{Motif Channel Opened in a White-Box:\\Stereo Matching via Motif Correlation Graph}
%\title{MatchAttention: Matching Relative Positions for High-Resolution Dense Correspondences}
\title{{\revisionon MatchAttention: Embedding Explicit Matching Constraints into Attention for Efficient Stereo Matching}}

\author{
	Tingman Yan\orcidlink{0000-0002-7370-6077},
	Tao Liu\orcidlink{0000-0003-1243-4546},~\IEEEmembership{Senior Member,~IEEE,}
	{\revisionon Chenghao Li\orcidlink{0000-0002-8968-7333},
	Quanli Liu\orcidlink{0000-0003-4118-2368},~\IEEEmembership{Member,~IEEE,}}
	Xilian Yang\orcidlink{0000-0002-2713-6392},
	Qunfei Zhao\orcidlink{0000-0002-9882-730X},
	Zeyang Xia\orcidlink{0000-0002-0075-7949},~\IEEEmembership{Senior Member,~IEEE}% <-this % stops a space
\IEEEcompsocitemizethanks{
	{\revisionon \IEEEcompsocthanksitem Tingman Yan, Tao Liu, Quanli Liu, and Xilian Yang are with the School of Control Science and Engineering, Dalian University of Technology, Dalian, 116024, China.
	E-mail: tingmanyan@dlut.edu.cn, tliu@dlut.edu.cn, liuql@dlut.edu.cn, yangxl@dlut.edu.cn}
	{\revisionon \IEEEcompsocthanksitem Chenghao Li is with Pika Labs. E-mail: lch94102@gmail.com}
	\IEEEcompsocthanksitem Qunfei Zhao is with the Department of Automation, Shanghai Jiao Tong University, Shanghai, 200240, China.
	E-mail: zhaoqf@sjtu.edu.cn
	\IEEEcompsocthanksitem Zeyang Xia is with the School of Mechanical Engineering, Shanghai Jiao Tong University, Shanghai, 200240, China.
	E-mail: zxia@sjtu.edu.cn
	\protect\\
	}% <-this % stops an unwanted space
	%\thanks{* Corresponding author: .}
	%\thanks{Manuscript received November 19, 2024}%; revised August 26, 2015.}
}

\IEEEtitleabstractindextext{%
\begin{abstract}
\revisionon Standard attention mechanisms are not well suited to stereo matching. Global attention scales quadratically and provides no explicit matching constraint, while local attention is efficient but loses long-range correspondences. We propose \textit{MatchAttention}, an attention mechanism that embeds an explicit matching constraint into attention by treating the relative position between a query and its matched key as a learnable component of attention sampling. Centering a small contiguous sampling window on this learnable relative position enforces the matching constraint and supports long-range correspondence at strictly linear attention complexity. A differentiable \textit{contiguous attention sampling} (CAS) operator enables sub-pixel accuracy, and cascaded MatchAttention blocks iteratively refine the relative positions through residual connections. We instantiate MatchAttention as a hierarchical coarse-to-fine stereo network with two variants. MatchAttentionXL targets accuracy and MatchAttentionRT targets real-time edge inference. MatchAttentionXL achieves state-of-the-art accuracy on Middlebury V3 and top results across KITTI 2012/2015 and ETH3D. MatchAttentionRT runs at 9.3\,ms on RTX 4060\,Ti and 79.1\,ms on Jetson Orin NX\,16\,GB at $1024\times512$, making it the first stereo model to deliver real-time edge inference without sacrificing zero-shot generalization. The code is available at \url{https://github.com/TingmanYan/MatchAttention}.
\end{abstract}

}

\maketitle

\IEEEdisplaynontitleabstractindextext

% \fi
\IEEEpeerreviewmaketitle

% Includegraphics override: keep images original (black) while text is blue
\ifdefined\TRACKEDCHANGES
\let\oldincludegraphics\includegraphics
\renewcommand{\includegraphics}[2][]{{\color{black}\oldincludegraphics[#1]{#2}}}
\fi

% Ensure body text is blue in tracked-changes mode
\revisionon

% Captions, tables, figures blue for tracked changes
\ifdefined\TRACKEDCHANGES
\let\oldcaption\caption
\renewcommand{\caption}[2][]{\oldcaption[#1]{{\color{revisionblue}#2}}}
\AtBeginEnvironment{table}{\color{revisionblue}}
\AtBeginEnvironment{table*}{\color{revisionblue}}
\AtBeginEnvironment{figure}{\color{revisionblue}}
\AtBeginEnvironment{figure*}{\color{revisionblue}}
\fi

% === Sections ===
\IEEEraisesectionheading{\section{Introduction}\label{sec:1:introduction}}

% Fig 1 declared at the very start of section 1 so LaTeX has the maximum
% opportunity to place it on page 1 (e.g. top of column 2).
\begin{figure}[!t]
\revisionon
	\centering
	\includegraphics[width=0.80\linewidth]{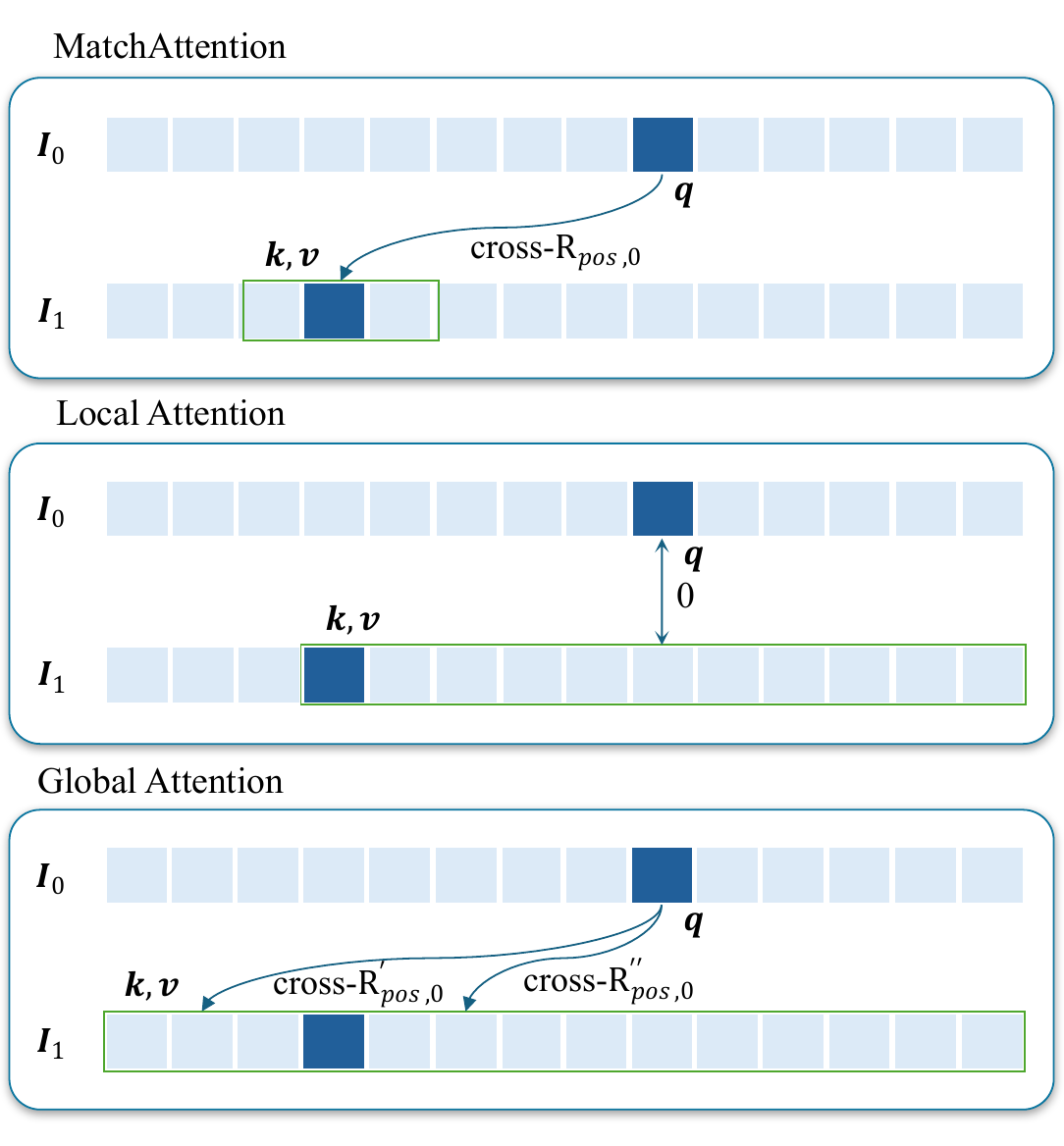}
	\caption{Comparison of local attention, global attention, and MatchAttention for stereo matching, illustrated in 1D.}
	\label{fig:1:attention_comparison}
\end{figure}

\revisionon
\IEEEPARstart{S}{tereo} matching, the recovery of a dense disparity field from a pair of rectified images, is a foundational problem in 3D computer vision. Accurate stereo serves as the depth backbone of autonomous driving, robot manipulation, and AR/VR applications \cite{stereopolicy, stereovla, stereonav, wang2023smartglasses}. It also provides a reliable geometric prior for 3D Gaussian splatting methods that derive scene geometry from stereo matching \cite{li2025bgsslam, han2024binocular3dgs, safadoust2024stereogs}. Across these use cases, accurate stereo must satisfy three requirements: (i) high accuracy, especially at the sub-pixel scale and on high-resolution inputs, where the depth error is inversely proportional to the disparity resolution; (ii) strong zero-shot generalization across domains, because dense ground-truth data for fine-tuning is rare in practice; and (iii) real-time inference on resource-constrained edge devices such as embedded GPUs. Each requirement can be met in isolation, but satisfying all three at once remains difficult.

Two architectural families currently dominate stereo matching. Cost-volume-based iterative refinement methods \cite{raftstereo, igev, selective, AIO, foundationstereo} build a correlation volume once from encoder features and iteratively refine disparity with convolutional GRUs \cite{raftstereo}, offering good accuracy but relying on a static, fixed correlation that does not adapt to the evolving correspondence during refinement. Attention-based methods \cite{sttr, xu2023unifying, croco} aggregate cross-view features through self- and cross-attention, naturally adapting features to the matching task, but inherit the quadratic complexity of global attention. Consequently, state-of-the-art accurate stereo networks are computationally expensive \cite{foundationstereo} (often requiring multi-second inference on high-end GPUs for a single high-resolution pair), and real-time variants \cite{hitnet, bangunharcana2021correlate, cre, igev++, liteanyStereo} typically trade zero-shot generalization for throughput.

We attribute this limitation to how existing attention mechanisms interact with the matching task. For a query in the reference view, the attention mechanism must simultaneously cover its true correspondence even when the disparity is large, distinguish that correspondence from the non-matching keys, and scale to high-resolution inputs. As illustrated in Fig.~\ref{fig:1:attention_comparison}, where cross-$R_{pos}$ denotes the cross-view relative position between a query and its matched key, the existing options each fail on at least one of the three requirements above. Local attention \cite{hassani2023neighborhood, Liu_2021_ICCV} is linear and discriminative but cannot cover large disparities. Global attention covers the match but is quadratic in complexity, and because every query attends to every key, the ground-truth match is always inside the attention range, so the network never receives the discriminative signal of missing the match, which degrades generalization in practice. Deformable attention \cite{zhu2021iclr} samples scattered, query-predicted offsets from single-view features without cross-view matching feedback, and the scattered sampling lacks the contiguous spatial support that dense correspondence requires.

For stereo matching, and more generally for dense correspondence, the relative position between a query and its true matched key is the learning target. The $x$-component of this vector is the disparity. Existing attention mechanisms do not exploit this connection. The relative position is typically treated as an auxiliary embedding (e.g., RoPE \cite{su2021roformer}) or predicted by a head added to the attention output, but not as a structural component of the attention operator itself.

Based on this observation, we introduce \textit{MatchAttention}, an attention mechanism that embeds the relative position directly into the attention sampling. Given a query, we compute its attention sampling window not at the query's own position (as in local attention), nor across the full image (as in global attention), but centered at the query position plus the currently predicted cross-$R_{pos}$, using a small contiguous window. This \emph{contiguous attention sampling} (CAS) imposes an explicit matching constraint, under which the ground-truth match falls inside the window only when cross-$R_{pos}$ is correct, so the network receives a clear gradient signal to refine cross-$R_{pos}$ toward the correct match. The differentiable CAS operator enables sub-pixel centering of the sampling window, and residual connections iteratively update the relative positions alongside the token features at every layer. By concatenating cross-$R_{pos}$ as a feature channel, self-MatchAttention propagates correspondence estimates among neighboring tokens, analogous to label propagation in belief propagation \cite{sun2003stereo}. Because the window is small (e.g., $4\times 4$ or even $1\times 4$) and the matching range is decoupled from the window size, MatchAttention is simultaneously long-range, discriminative, and strictly linear in token count.

We instantiate MatchAttention in a hierarchical coarse-to-fine stereo network that interleaves self- and cross-MatchAttention blocks across scales from $1/32$ to $1/4$, with two variants: MatchAttentionXL for high-accuracy evaluation and MatchAttentionRT for real-time deployment. To handle cross-view occlusions, we propose a gated cross-MatchAttention module and a consistency-constrained loss that together suppress occluded regions in both forward and backward passes. Although the non-parameterized modules of MatchAttention account for only $\approx$3--5\% of the FLOPs, they dominate the wall-clock latency ($> 70\%$) on GPUs because of the random-access \texttt{gather} operations driven by the dynamically updated relative positions. We therefore introduce three hardware-aware optimizations: channel compression, a unified cross-scale window, and horizontal-only matching, which together yield a $\sim$3$\times$ end-to-end speedup with minimal accuracy drop.

MatchAttentionXL ranks among the top-3 published methods on most Middlebury V3 error metrics and attains top-tier performance on KITTI 2012, KITTI 2015, and ETH3D. MatchAttentionRT runs at 9.3\,ms on an RTX 4060\,Ti, 36.7\,ms on a Jetson Orin AGX 64\,GB, and 79.1\,ms on a Jetson Orin NX 16\,GB at $1024\times 512$ resolution. To the best of our knowledge, MatchAttentionRT is the first stereo model to achieve real-time inference on edge devices while retaining zero-shot generalization competitive with recent non-real-time methods, surpassing ZeroStereo~\cite{zerostereo} on most evaluated metrics.

Our main contributions are as follows.
\begin{itemize}
    \item We propose MatchAttention by identifying the core limitation of standard attention in stereo matching. Neither local, global, nor deformable attention embeds an \emph{explicit, learnable matching constraint} in the operator itself. MatchAttention addresses this limitation by treating the relative position between a query and its matched key as the learning target for dense correspondence and by centering a small contiguous sampling window on learnable relative positions through a differentiable contiguous attention sampling (CAS) operator.
    \item We identify random-access \texttt{gather} as the practical latency bottleneck of MatchAttention and propose three hardware-aware optimizations that bridge theoretical linear complexity and real-time inference on embedded GPUs.
    \item We propose MatchAttentionXL and MatchAttentionRT, two hierarchical stereo networks based on MatchAttention with occlusion handling in forward and backward passes. Extensive experiments on public benchmarks and embedded devices demonstrate the accuracy of MatchAttentionXL and the real-time zero-shot performance of MatchAttentionRT.
\end{itemize}

\revisionon
\section{Related Work}
\label{sec:2:related}
\subsection{Attention Mechanisms}

Originally proposed for neural machine translation \cite{Bahdanau2015}, the attention mechanism captures global dependencies by computing all-pair correlations followed by a softmax activation. The quadratic complexity of global attention has motivated a range of efficient variants, each balancing the matching range, the sampling structure, and the computational cost.

\noindent\textbf{Sparse attention} reduces computational overhead by exploiting sparsity within the attention matrix. For instance, Longformer \cite{Beltagy2020Longformer} combines dilated sliding-window attention with global tokens to achieve linear complexity, Big Bird \cite{zaheer2020bigbird} augments this approach with random attention, and SparseK attention \cite{lou2024sparser} selects the top-$k$ key-value pairs via differentiable scoring. Although effective for natural language processing, these unstructured sparsity patterns do not capture the geometric constraints inherent in visual correspondence.

\noindent\textbf{Sliding-window attention} restricts each query to a fixed-size local window, thereby ensuring linear complexity. Notable examples include Neighborhood Attention (NAT) \cite{hassani2023neighborhood}, which attends to nearest-neighbor tokens; CLEAR \cite{liu2024clear}, which introduces a circular design for diffusion transformers; and the Swin Transformer \cite{Liu_2021_ICCV}, which partitions tokens into non-overlapping sub-windows connected by shifted operations. Because the window center remains aligned with the query, however, the matching range is coupled to the window size. This limitation is significant for stereo matching, where large disparities are common in close-range or wide-baseline settings.

\noindent\textbf{Deformable attention} \cite{zhu2021iclr} overcomes local range limitations by predicting data-dependent offsets from the query, allowing it to attend to a sparse set of scattered tokens. This paradigm has been widely adopted in object detection and bird's-eye-view (BEV) feature aggregation \cite{li2022bev}. While earlier variants, such as hard attention \cite{xu2015icml} and local-predictive attention \cite{luong-etal-2015-effective}, suffered from non-differentiability due to discrete sampling, modern deformable attention achieves a flexible receptive field with linear cost. However, its sampling offsets are predicted from single-view query features without cross-view geometric feedback. The resulting scattered sampling points lack the contiguous spatial structure necessary for reliable dense correspondence.

Positional information is typically supplied to attention as an auxiliary signal, through fixed sinusoidal encodings, learnable relative embeddings \cite{Liu_2021_ICCV, Liu_2022_CVPR, shi2023transnext}, and rotary positional embedding (RoPE) \cite{su2021roformer}, the last of which is widely adopted in modern foundation models \cite{grattafiori2024llama3herdmodels, croco, dust3r_cvpr24}. These embeddings inject spatial priors into the attention weights but remain \emph{decoupled} from the sampling locations themselves.

In summary, existing efficient attention mechanisms are not well suited to stereo matching. Sparse attention ignores geometric structure, sliding-window attention couples the receptive field to the window size, deformable attention lacks cross-view feedback and contiguous sampling, and positional embeddings act only as auxiliary signals. None of them simultaneously provides long-range matching, explicit correspondence modeling, and linear complexity. MatchAttention addresses these limitations by treating relative position as both the sampling coordinate and the explicit learning target, centering a contiguous window on the predicted cross-view correspondence.

\subsection{Accurate Stereo Matching}
\label{sec:2.2:accurate}
High-resolution, high-accuracy stereo matching has steadily improved over the past decade through cost-volume filtering, iterative refinement, and transformer-based aggregation. Early end-to-end methods \cite{gcnet, psmnet, bangunharcana2021correlate} construct and regularize 4D cost volumes using 3D convolutional networks, but exhibit limited cross-domain generalization and remain prohibitively expensive at high resolutions.

To circumvent this bottleneck, RAFT-Stereo \cite{raftstereo} introduced an iterative refinement paradigm in which a global correlation volume is computed once via matrix multiplication and subsequently refined with lightweight 2D convolutions. This framework became the first deep learning method to surpass traditional optimization-based approaches \cite{Taniai18, yan2023tip} on the Middlebury benchmark and has since spawned a rich family of extensions. IGEV-Stereo \cite{igev} introduces geometry encoding volumes for iterative updates, and Selective-Stereo \cite{selective} fuses multi-frequency GRU branches to recover fine details. CREStereo \cite{cre} employs cascaded recurrent networks with adaptive correlation computation. AIO-Stereo \cite{AIO} introduces adaptive distillation from vision foundation models \cite{oquab2024dinov, Kirillov_2023_ICCV, dav2}. FoundationStereo \cite{foundationstereo} builds a zero-shot stereo foundation pipeline with large-scale synthetic data, side-tuning from monocular depth priors \cite{dav2}, and attentive hybrid cost filtering. DEFOM-Stereo \cite{defom_stereo} integrates depth-foundation-model cues into recurrent stereo with a scale-update module for metric disparity recovery. MonSter++ \cite{monster++} improves recurrent stereo by injecting monocular depth priors into a coarse-to-fine update process. StereoAnywhere \cite{stereo_anywhere} combines stereo geometry and monocular depth foundation priors through a dual-branch cost-volume fusion design to improve robustness in challenging regions such as textureless and non-Lambertian areas. ZeroStereo \cite{zerostereo} synthesizes stereo pairs from single images by first warping the input image with monocular depth, then using diffusion-based inpainting to complete the occluded regions. A central limitation shared by many cost-volume-based methods is that their correlation volumes are built from initial encoder features and remain static during refinement, which limits dynamic adaptation as correspondence evolves.

Transformer-based methods address this limitation with dynamic, context-aware cross-view feature interactions. STTR \cite{sttr} frames stereo matching as sequence-to-sequence translation under optimal transport; UniMatch \cite{xu2023unifying} unifies stereo, flow, and depth through interleaved self- and cross-attention; and CroCo-Stereo \cite{croco} employs a ViT encoder--decoder pre-trained via cross-view completion, achieving competitive accuracy without cost volumes or iterative refinement. S2M2 \cite{s2m2} pairs a multi-resolution global matching transformer with global and local refinement, using a loss that concentrates probability on feasible matches to jointly estimate disparity, occlusion, and confidence at state-of-the-art accuracy. While these models benefit from joint optimization of features and matching, their reliance on global cross-attention imposes quadratic complexity and typically requires tiling at high resolution \cite{croco}, which limits deployment on resource-constrained platforms.

MatchAttention avoids both limitations by embedding the matching target directly inside a linear-complexity attention operator. It performs iterative correspondence refinement and dynamic feature aggregation simultaneously, circumventing the static-correlation bottleneck of cost-volume methods and the quadratic cost of global transformers.

\subsection{Efficient Stereo Matching}
\label{sec:2.3:efficient}
Driven by autonomous driving, robotics, and AR/VR, a parallel line of work targets stereo matching at real-time rates, especially on embedded devices. HITNet \cite{hitnet} combines hierarchical tile-based initialization with slanted-plane refinement for high throughput, and CoEx \cite{bangunharcana2021correlate} guides cost-volume excitation with image features, substantially reducing 3D-CNN cost. RT-IGEV++ \cite{igev++} derives a real-time variant from IGEV++ using a simplified geometry volume and a lighter recurrent update module. Fast-FSD \cite{fastfoundationstereo} accelerates FoundationStereo through distillation, blockwise architecture search, and structured pruning. RT-Monster++ \cite{monster++} achieves real-time speed through a coarse-to-fine framework with local cost volumes and cascaded disparity search. LightStereo \cite{lightstereo} uses channel-boosted 3D cost aggregation, and LiteAnyStereo \cite{liteanyStereo} combines a compact backbone, hybrid cost aggregation, and a three-stage million-scale training strategy to improve zero-shot efficiency. BANet \cite{banet} decomposes the cost volume into detailed and smooth branches for bilateral aggregation with mobile-friendly computation. IINet \cite{iinet} replaces heavy explicit 3D aggregation with a compact implicit 2D intra-inter fusion design. Pip-Stereo \cite{pipstereo} progressively prunes recurrent iterations toward single-pass inference. MatchAttentionRT follows a different approach: rather than relying on distillation, pruning, or architecture search, it achieves real-time edge inference directly from the operator, whose explicit matching constraint and linear complexity make it efficient without these additional steps.

\subsection{Relation to Existing Methods}
\label{sec:2:relation}
Existing stereo approaches can be grouped by how they formulate correlation and attention. \textit{Correlation-based methods} \cite{raftstereo, pwc_net, igev, selective, foundationstereo} explicitly build a cost volume and regress the final disparity with a separate refinement decoder. The matching representation is structurally decoupled from feature learning, and the cost volume remains static during refinement. \textit{Attention-based methods} \cite{xu2023unifying, xu2022gmflow, croco, sttr} use cross-attention to aggregate features across views, supporting dynamic updates, but their reliance on global attention incurs $\mathcal{O}(N^2)$ cost that limits scalability to high resolution.

MatchAttention unifies these two paradigms. It performs explicit spatial matching by computing query--key similarities strictly within a localized, constant-sized window of $(w{+}1)^2$ tokens, while simultaneously aggregating cross-view contextual features. Critically, this sampling window is dynamically centered on the evolving cross-$R_{pos}$ (and shifted by self-$R_{pos}$ within each view), so the matching range is decoupled from the window size, and the complexity is strictly $\mathcal{O}(N)$. Because features and relative positions are jointly updated at every layer, MatchAttention combines the iterative refinement of correlation-based stereo architectures with the data-driven feature learning of modern transformers. Unlike recent dataset-scaling approaches \cite{liteanyStereo, zerostereo}, our improvements are at the operator level and are therefore complementary to advances on the data and training side.

\revisionon
\begin{figure*}[!t]
\revisionon
	\centering
	\includegraphics[width=0.98\linewidth]{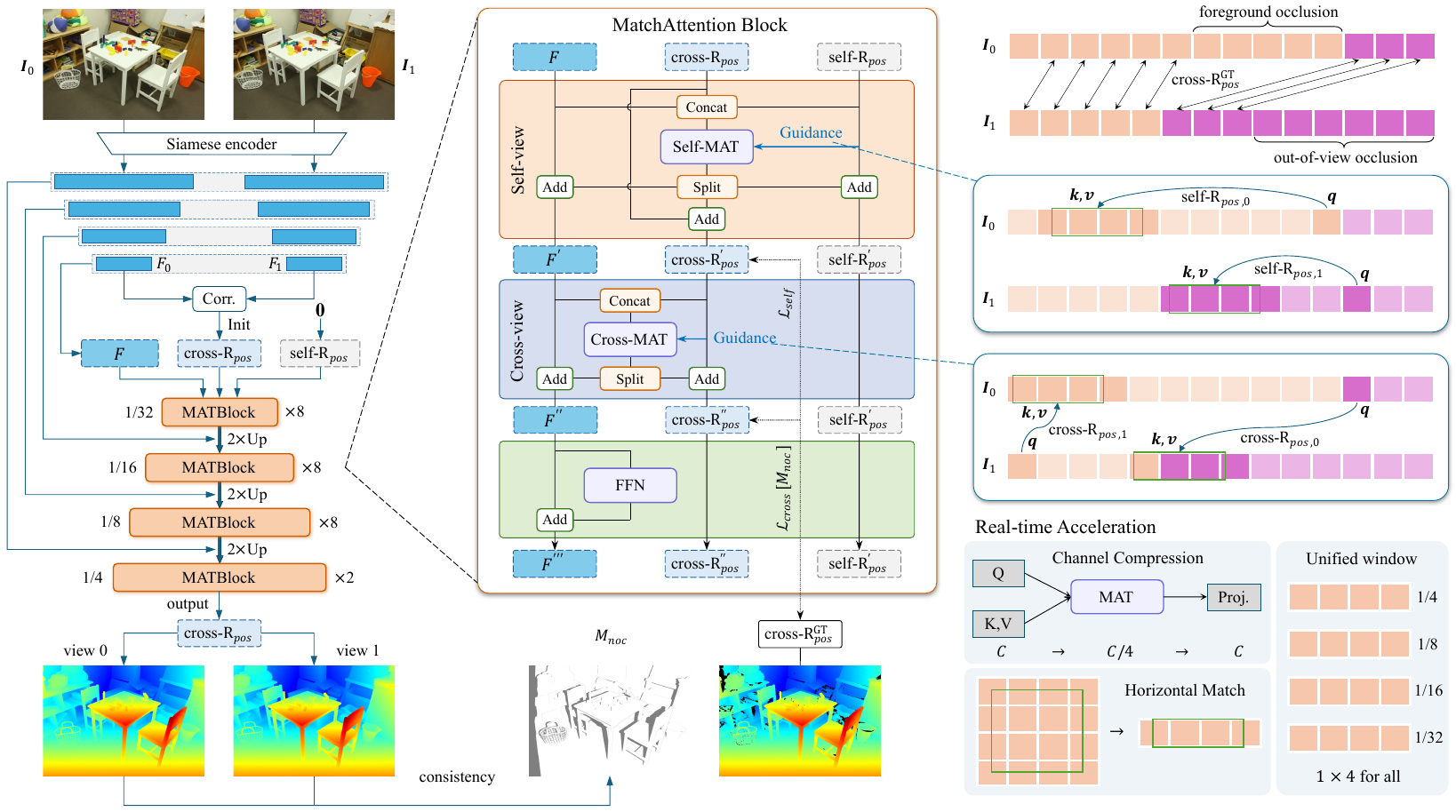}
	\caption{Overview of the proposed MatchAttention stereo architecture. (Left) Hierarchical encoder--decoder pipeline. (Middle) MatchAttention block (MATBlock). (Right) 1D illustration of how self-$R_{pos}$ and cross-$R_{pos}$ control contiguous attention sampling (CAS). (Bottom-right) Hardware-aware acceleration.}
	\label{fig:2:overview}
\end{figure*}

\section{Methodology}
\label{sec:3:methodology}

\subsection{Overview}
\label{sec:3.0:overview}

Figure~\ref{fig:2:overview} presents the full architecture. Given left and right images $I_0$ and $I_1$, a shared-weight siamese CNN encoder \cite{yu2024metaformer, hendrycks2023gaussianerrorlinearunits} extracts a feature pyramid at $\{1/4, 1/8, 1/16, 1/32\}$ scales. At the coarsest scale ($1/32$), a feature correlation along the epipolar line initializes the cross-view relative position cross-$R_{pos}$, while the self-view relative position self-$R_{pos}$ is initialized to zero. These two fields, together with the encoder features, are fed into a sequence of \emph{MatchAttention Blocks} (MATBlocks) that refine them progressively from $1/32$ to $1/4$ scale. Between scales, features are upsampled with transposed convolution \cite{Ronneberger2015unet} and both cross-$R_{pos}$ and self-$R_{pos}$ are upsampled with the convex upsample module \cite{raft}. Finally, cross-$R_{pos}$ is convex-upsampled to full resolution. Its $x$-component is directly the stereo disparity.

Each MATBlock (Fig.~\ref{fig:2:overview}, middle) refines $(F, \mathrm{cross}\text{-}R_{pos}, \mathrm{self}\text{-}R_{pos})$ through self-MatchAttention, cross-MatchAttention, and an FFN, with each stage applying LayerNorm and a residual connection (detailed in Section~\ref{sec:3.2:block}).

Fig.~\ref{fig:2:overview} (right) illustrates the matching mechanism in 1D. For self-MatchAttention, self-$R_{pos}$ shifts the sampling window in occluded regions toward non-occluded neighbors, allowing them to aggregate valid context. For cross-MatchAttention, cross-$R_{pos}$ centers a small CAS window around the putative match. The window is deliberately small so that the network can distinguish correct from incorrect matches, which is key to accurate matching and strong generalization (Section~\ref{sec:3.1:matchattention}).

The network is supervised on both intermediate fields produced inside each MATBlock, namely cross-$R_{pos}'$ from self-MatchAttention and cross-$R_{pos}''$ from cross-MatchAttention. A non-occlusion mask $M_{noc}$ excludes occluded regions from the cross-$R_{pos}''$ supervision (Section~\ref{sec:3.3:occlusion}). The real-time acceleration path (Fig.~\ref{fig:2:overview}, bottom-right) closes the gather-bandwidth gap through three hardware-aware optimizations (Section~\ref{sec:3.4:acceleration}).

The following subsections detail the MatchAttention mechanism (Sec.~\ref{sec:3.1:matchattention}), the MATBlock design (Sec.~\ref{sec:3.2:block}), occlusion handling (Sec.~\ref{sec:3.3:occlusion}), and hardware-aware acceleration (Sec.~\ref{sec:3.4:acceleration}).

\subsection{The MatchAttention Mechanism}
\label{sec:3.1:matchattention}

\subsubsection{Motivation and Formulation}

A query token $\mathbf{q}_i$ from the reference view has exactly one matched key $\mathbf{k}_{i^*}$ in non-occluded regions of the target view, and the offset between their pixel positions is the disparity. We therefore designate the offset between the query and the center of its key/value sampling window as a learnable \emph{relative position} $\mathbf{r}_i = \mathrm{cross}\text{-}R_{pos}[i,:]$ and center the sampling window at $\mathbf{p}_i^k = \mathbf{p}_i^q + \mathbf{r}_i$. Because $\mathbf{p}_i^k$ is continuous, we floor it to $\lfloor \mathbf{p}_i^k \rfloor$ and define an expanded discrete sampling window $\overline{\mathcal{W}}_i$ of size $(w+1)^2$.

Given the projected query $Q = W_q\hat{F}_0$, key $K = W_k\hat{F}_1$, and value $V = W_v\hat{F}_1$, the MatchAttention output for the $i$-th query is computed as
\begin{gather}
	\label{eq:3:matchattention}
	\begin{aligned}
		Match&Attention_{w} (Q, K, V, \mathrm{cross}\text{-}R_{pos}) [i] = W_p\, \mathbf{m}_i \\
		&= W_p \sum_{j \in \overline{\mathcal{W}}_i} \tilde{\sigma}(\langle \mathbf{q}_i, \mathbf{k}_j \rangle)\, \mathbf{v}_j,
	\end{aligned}
\end{gather}
where $\overline{\mathcal{W}}_i = \{ \lfloor \mathbf{p}_i^k \rfloor + (u,v) \mid u,v = -w/2,\ldots,w/2+1 \}$, $\tilde{\sigma}$ is the differentiable CAS operator (Section~\ref{sec:3.1:cas}), and $W_p$ is the output projection. In cross-MatchAttention, $\hat{F}_0$ and $\hat{F}_1$ come from different views, and both cross-view directions (left-to-right and right-to-left) are computed simultaneously. In self-MatchAttention, both come from the same view and $\mathrm{cross}\text{-}R_{pos}$ is replaced by $\mathrm{self}\text{-}R_{pos}$ in Eq.~\ref{eq:3:matchattention}.

\subsubsection{Explicit Matching Constraint}

In MatchAttention, $\mathrm{cross}\text{-}R_{pos}$ constrains attention sampling to a small window around the putative match. The attention output in turn predicts a refined $\mathrm{cross}\text{-}R_{pos}$. An incorrect $\mathrm{cross}\text{-}R_{pos}$ excludes the true match from the sampled support, increasing the supervised correspondence loss and shaping gradients through the differentiable CAS operator. Figure~\ref{fig:1:attention_comparison} contrasts this feedback loop with local and global attention.

When $\mathrm{cross}\text{-}R_{pos} \!=\! 0$, the window is pinned to the query and MatchAttention reduces to \emph{local attention}. Covering large disparities now requires a large window, which increases computational cost and disperses the attention signal. When the window spans the entire image, MatchAttention reduces to \emph{global attention}. The ground-truth match is always inside the window regardless of $\mathrm{cross}\text{-}R_{pos}$, so the network never receives a penalty for a wrong prediction. Removing this signal degrades zero-shot generalization (Section~\ref{sec:4.4:zero_shot}).

MatchAttention avoids both failure modes. With a small window and a learnable $\mathrm{cross}\text{-}R_{pos}$, the ground-truth match falls inside the window only when $\mathrm{cross}\text{-}R_{pos}$ is correct. An incorrect $\mathrm{cross}\text{-}R_{pos}$ excludes the true match from the sampled support, which increases the supervised correspondence loss. Because CAS is differentiable with respect to the sampling location, this loss provides gradients that refine $\mathrm{cross}\text{-}R_{pos}$ toward the correct correspondence. This explicit matching constraint is the central property of MatchAttention. A smaller window produces a sharper penalty and stronger generalization, as confirmed by the ablation study (Section~\ref{sec:4.5:ablation}).

\subsubsection{CAS: Sub-Pixel, Differentiable Sampling}
\label{sec:3.1:cas}

\begin{figure}[!t]
\revisionon
	\centering
	\includegraphics[width=0.95\linewidth]{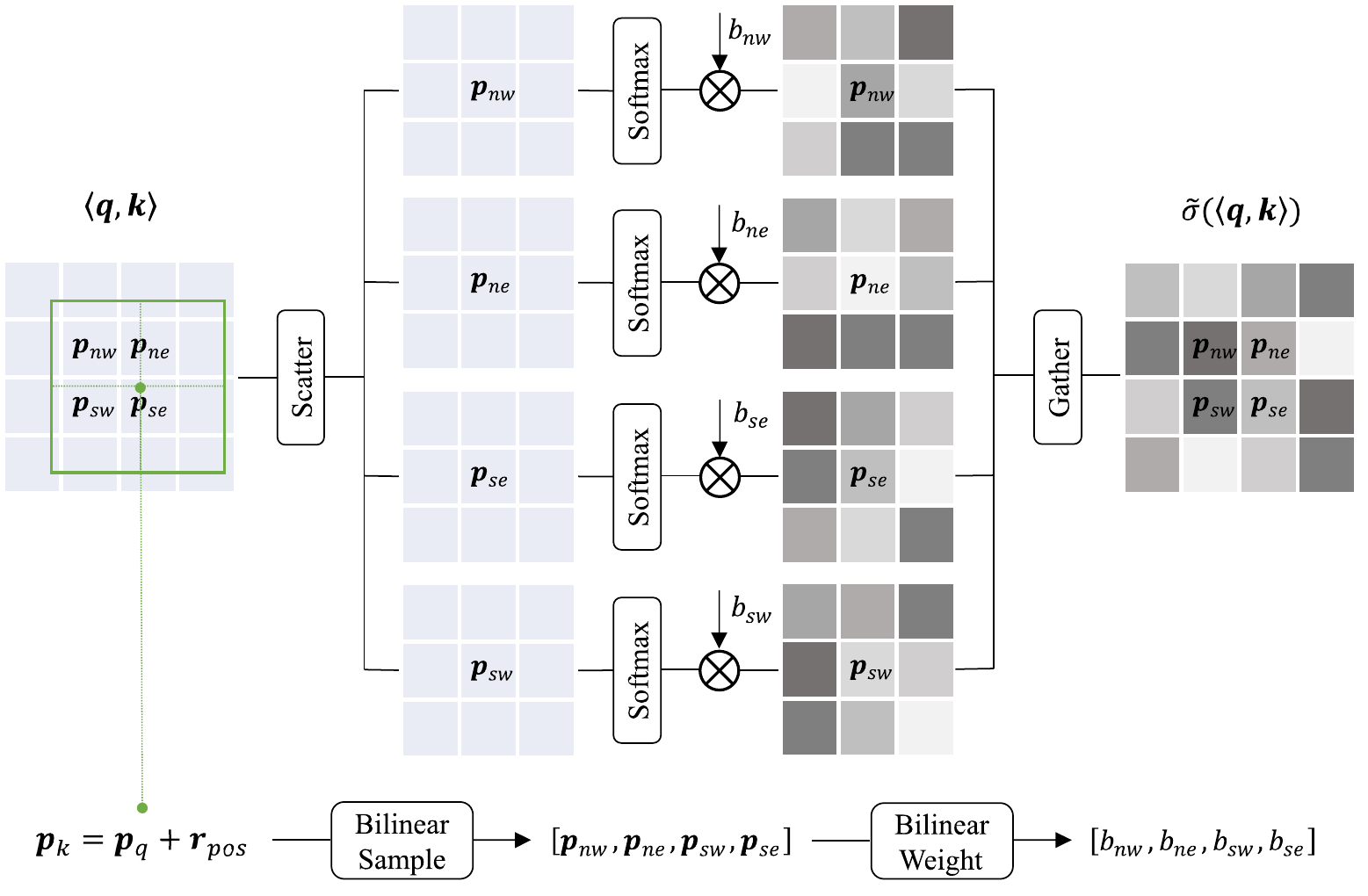}
	\caption{Illustration of contiguous attention sampling (CAS). The continuous key position is split into four sub-windows for sub-pixel attention sampling.}
	\label{fig:3:cas}
\end{figure}

Because $\mathbf{p}_i^k$ is a continuous coordinate while the key tokens are defined on a discrete grid, we propose \emph{contiguous attention sampling} (CAS, $\tilde{\sigma}$) to perform sub-pixel window-level sampling. The four nearest integer positions $\mathbf{p}_i^{nw}, \mathbf{p}_i^{ne}, \mathbf{p}_i^{sw}, \mathbf{p}_i^{se}$ serve as sub-window centers (Fig.~\ref{fig:3:cas}). Each sub-window $\mathcal{W}_i^t$ ($t \!\in\! \mathcal{T} \!=\! \{nw,ne,sw,se\}$) of size $w^2$ is scattered from $\overline{\mathcal{W}}_i$. Softmax is computed independently within each sub-window, scaled by the corresponding bilinear weight $b_i^t$, and gathered back to $\overline{\mathcal{W}}_i$:
\begin{gather}
	\label{eq:3:cas}
	\bar{\alpha}_{ij} = \tilde{\sigma}(\langle \mathbf{q}_i, \mathbf{k}_j \rangle) = \sum_{t \in \mathcal{T}} \frac{b_i^t}{Z_i^t}\, \exp(\langle \mathbf{q}_i, \mathbf{k}_{j^t} \rangle),
\end{gather}
where $j^t \!\in\! \mathcal{W}_i^t \!\subset\! \overline{\mathcal{W}}_i$. CAS is fully differentiable: gradients with respect to cross- or self-$R_{pos}$ flow through both the residual connection and the bilinear weights $\{b_i^t\}$. A key computational advantage is that both the query-key similarity and the value aggregation are evaluated on the discrete $(w{+}1)^2$ expanded grid, whereas naive bilinear sampling of keys and values would require $4 w^2$ operations per query. By exploiting the overlap among the four sub-windows, CAS visits each query-key pair exactly once, which reduces computation by approximately $4\times$. The formulation above describes the 2D case. For horizontal stereo matching, CAS reduces to a 1D variant with two sub-windows (left and right), further simplifying computation (Section~\ref{sec:3.4:acceleration}).

\subsubsection{Negative $L_1$-norm Similarity}

Because each reference pixel corresponds to at most one true match, dense correspondence is intrinsically sparse. To exploit this sparsity, we replace the standard dot-product similarity with the \emph{negative $L_1$-norm}:
\begin{gather}
	\label{eq:3:l1norm}
	\mathrm{Softmax}(\langle \mathbf{q}_i, \mathbf{k}_j \rangle) = \frac{1}{Z_i}\, \exp(-\gamma \| \mathbf{q}_i - \mathbf{k}_j \|_1), \quad \gamma = 1/\sqrt{c_k}.
\end{gather}
The combination of softmax and the negative $L_1$ acts as a normalized Laplace kernel. It sharpens the attention distribution and improves edge and detail recovery.

\subsubsection{Relative-Position Updates and Multi-Head Extension}

The relative positions are concatenated with the token features and updated layer by layer through residual connections. Accordingly, the weight matrices are expanded by two channels, yielding $W_q \!\in\! \mathbb{R}^{(c_{in}+2)\times c_k}$, $W_k \!\in\! \mathbb{R}^{(c_{in}+2)\times c_k}$, $W_v \!\in\! \mathbb{R}^{(c_{in}+2)\times c_v}$, and $W_p \!\in\! \mathbb{R}^{c_v \times (c_{out}+2)}$.

Multi-head attention is implemented in the standard way~\cite{Vaswani2017attention} where the channel dimension is split into $h$ heads, each operating on a $c/h$-dimensional subspace, and their outputs are concatenated before the output projection. In self-MatchAttention, each head maintains its own self-$R_{pos}$, allowing the model to jointly attend to multiple non-occluded neighborhoods. In cross-MatchAttention, all heads share a single cross-$R_{pos}$, reflecting the one-to-one nature of stereo correspondence.

\subsubsection{Complexity Analysis}

MatchAttention contains two kinds of operators. The \emph{parameterized} linear projections have strictly linear cost and run on tensor cores. The \emph{non-parameterized} operators (query-key similarity, CAS, and value aggregation) run on general-purpose CUDA cores, because the per-query dynamics of cross-$R_{pos}$ and self-$R_{pos}$ prevent tensor-core batching. For query, key, and value tokens of spatial resolution $H\!\times\!W$, these three operators require $2HWh c_k(w+1)^2$, $HWh[20 + 2(w+1)^2 + 12w^2]$, and $2HWh c_v(w+1)^2$ FLOPs, respectively. Since $w$ is a small constant, the overall complexity is $\mathcal{O}(HWh\max(c_k, c_v) w^2)$, strictly linear in the number of tokens $HW$. The intermediate attention tensor occupies $HWh(w+1)^2$ memory, which is also linear and stays smaller than the token features whenever $(w+1)^2 < \max(c_k, c_v)$.

\subsection{MatchAttention Block}
\label{sec:3.2:block}

The MatchAttention Block (Fig.~\ref{fig:2:overview}, middle) is the repeat unit of the decoder. Each block refines $(F, \mathrm{cross}\text{-}R_{pos}, \mathrm{self}\text{-}R_{pos})$ through three sub-modules, each applied with LayerNorm and a residual connection. All three tensors carry both views, e.g., $F = F^l \Vert F^r$.

\noindent\textbf{Self-MatchAttention.} The input is $\hat{F} = F \Vert \mathrm{cross}\text{-}R_{pos} \Vert \mathrm{self}\text{-}R_{pos}$ within each view. Self-$R_{pos}$ determines the CAS window and is updated by the residual connection. Because cross-$R_{pos}$ is concatenated as a feature channel, the attended tokens propagate their correspondence estimates to neighbors, an effect that shares the intuition of label propagation in traditional belief propagation~\cite{sun2003stereo, yan2023tip}. The intermediate cross-$R_{pos}'$ is supervised with an unmasked $L_1$ across all regions (Section~\ref{sec:3.3:occlusion}).

\noindent\textbf{Cross-MatchAttention.} The input is $\hat{F} = F' \Vert \mathrm{cross}\text{-}R_{pos}'$, with cross-$R_{pos}'$ centering the CAS window across views. The flattened attention weights $\bar{\boldsymbol{\alpha}}_i \!=\! \{\bar{\alpha}_{ij} \mid j \!\in\! \overline{\mathcal{W}}_i\}$ are appended to the projection input, so Eq.~\ref{eq:3:matchattention} becomes
\begin{gather}
	\label{eq:3:cross_matchattn}
	W'_p\, (\mathbf{m}_i \Vert \bar{\boldsymbol{\alpha}}_i),\quad W'_p \in \mathbb{R}^{(c_v + (w+1)^2) \times (c_{out}+2)},
\end{gather}
where $\bar{\boldsymbol{\alpha}}_i \!\in\! \mathbb{R}^{(w+1)^2}$ acts as an explicit local matching cue. The residual update yields cross-$R_{pos}''$, supervised with an $M_{noc}$-masked $L_1$ that excludes occluded regions (Section~\ref{sec:3.3:occlusion}).

\noindent\textbf{FFN (ConvGLU).} Each MATBlock closes with a depth-wise-convolutional gated linear unit~\cite{shi2023transnext} that refines $F$. Its gate suppresses unreliable aggregated features, and its depth-wise convolution serves as an implicit conditional positional embedding~\cite{chu2023conditional}.

\subsection{Occlusion Handling}
\label{sec:3.3:occlusion}

Enforcing supervision on cross-view occlusions would match occluded reference pixels to the foreground in the other view, while their features would contaminate the attention aggregation. We suppress them in the forward pass with a learned gate, and in the backward pass with a consistency-constrained loss that computes an online non-occlusion mask $M_{noc}$ without occlusion labels.

\noindent\textbf{Gate (forward pass).} The aggregated cross-view feature of Eq.~\ref{eq:3:cross_matchattn} is modulated with a learned gate $G \!=\! \mathrm{SiLU}(W_g \hat{F}_0)$ computed from reference-view features alone:
\begin{gather}
	\label{eq:3:gated_cross_matchattn}
	W'_p\, \big( (\mathbf{g}_i \odot \mathbf{m}_i) \Vert \bar{\boldsymbol{\alpha}}_i \big),\quad \mathbf{g}_i \!=\! G[i,:],
\end{gather}
with $W_g \!\in\! \mathbb{R}^{(c_{in}+2) \times c_k}$. The gate suppresses unreliable cross-view content in occluded or low-texture regions. Self-MatchAttention, concurrently, continues to refine cross-$R_{pos}$ inside those regions by borrowing monocular context from neighboring non-occluded tokens.

\noindent\textbf{Mask (backward pass).} Because cross-$R_{pos}$ is estimated jointly for both views, an online non-occlusion mask is built per layer during training from a cross-view consistency check:
\begin{gather}
	\label{eq:3:consistency_check}
	\begin{aligned}
		\mathrm{cross}\text{-}R_{pos,0}^{1\rightarrow 0} &= \mathcal{B}(\mathrm{cross}\text{-}R_{pos,1},\,\mathrm{cross}\text{-}R_{pos,0} + G_{2D}), \\
		M_{noc,0} &= \big\| \mathrm{cross}\text{-}R_{pos,0} + \mathrm{cross}\text{-}R_{pos,0}^{1\rightarrow 0} \big\|_1 \leq A,
	\end{aligned}
\end{gather}
with $\mathcal{B}$ bilinear sampling, $G_{2D}$ the coordinate grid, $A$ a tolerance threshold, and $M_{noc,1}$ obtained symmetrically. The cross-MatchAttention branch is supervised with an $M_{noc}$-masked $L_1$ augmented by a bidirectional consistency term:
\begin{gather}
	\label{eq:3:loss_cross}
	\begin{aligned}
		L_{cross}^l = &\big\| (\mathrm{cross}\text{-}R_{pos,0}^{gt} - \mathrm{cross}\text{-}R_{pos,0}^{l,\mathrm{cross}}) \cdot M_{noc,0}^l \big\|_1 \\
		+ &\epsilon\, \big\| (\mathrm{cross}\text{-}R_{pos,0} + \mathrm{cross}\text{-}R_{pos,0}^{1\rightarrow 0}) \cdot M_{noc,0}^l \big\|_1.
	\end{aligned}
\end{gather}
The self-MatchAttention branch instead uses an \emph{unmasked} $L_1$ on cross-$R_{pos}'$, propagating supervision into occluded regions. The initial cross-$R_{pos}$ is supervised with a multi-modal cross-entropy on the $1/32$-scale cost volume~\cite{Xu_CVPR_2024_ADL}. All per-layer sub-losses are combined with the exponentially-increasing weights standard in iterative refinement for stereo~\cite{raftstereo}.

\begin{figure}[!t]
\revisionon
	\centering
	\includegraphics[width=0.95\linewidth]{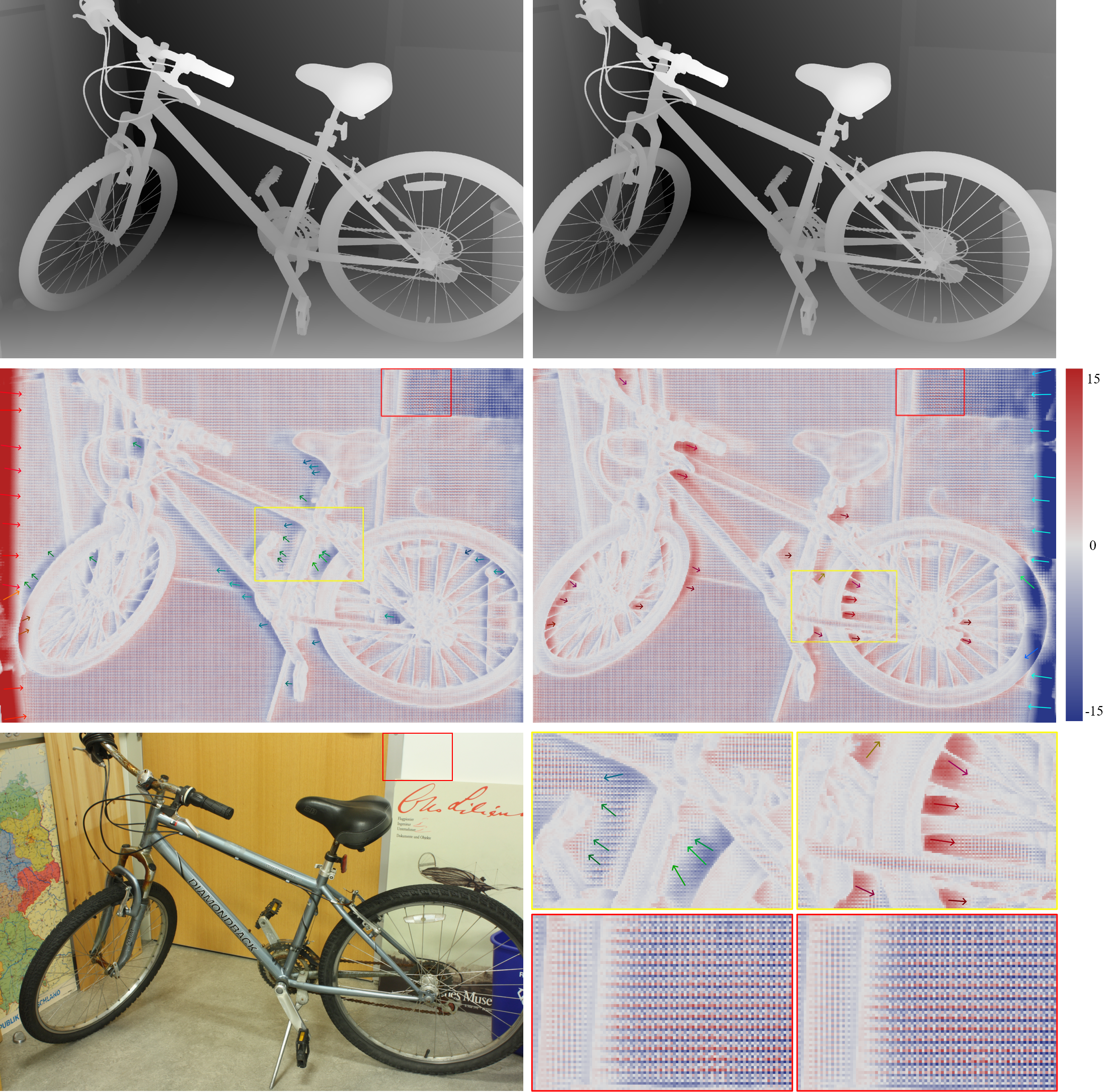}
	\caption{Visualization of learned self-$R_{pos}$ on Middlebury-F (Bicycle), zero-shot MatchAttentionXL (FSD~Mix). Red-to-blue encodes self-$R_{pos}\text{-}x$, and each arrow points from a query location to its sampling target. In out-of-view and foreground-occlusion regions, arrows point to nearby non-occluded areas. In textureless regions, self-$R_{pos}\text{-}x$ shows an alternating positive/negative pattern.}
	\label{fig:4:self_rpos}
\end{figure}

\noindent\textbf{Emergent Interpretability.} Figure~\ref{fig:4:self_rpos} visualizes learned self-$R_{pos}$ in self-MatchAttention under zero-shot evaluation. In out-of-view occluded regions (left side of the left view and right side of the right view), self-$R_{pos}$ vectors consistently point toward nearby non-occluded neighbors. In foreground-occlusion regions (zoomed view), they similarly point to the nearest non-occluded neighbors. In textureless regions, self-$R_{pos}\text{-}x$ exhibits an alternating positive/negative pattern, while near edges it remains close to zero, providing matching cues where appearance is ambiguous. This behavior suggests that self-MatchAttention learns an occlusion-aware and texture-aware sampling strategy from data, which is also consistent with the supplementary qualitative comparisons on large textureless surfaces (Section~S3).

\subsection{Real-Time Acceleration}
\label{sec:3.4:acceleration}

MatchAttention is theoretically linear, but its practical latency on GPUs is dominated by a different bottleneck. On a single RTX 4060 Ti, at $1280\!\times\!736$ resolution, the initial unaccelerated configuration in Fig.~\ref{fig:5:acceleration_cascade} spends $> 70\%$ of its wall-clock time inside MatchAttention, even though MatchAttention contributes only $\sim$3--5\% of the FLOPs. The cause is the random-access \texttt{gather} implicit in key/value sampling. Because cross-$R_{pos}$ and self-$R_{pos}$ differ per query and are updated at every layer, memory addresses are irregular and arithmetic intensity is low. Tensor cores then remain underutilized and the kernel becomes memory-bandwidth-bound.

To close this gap, three optimizations are introduced (Fig.~\ref{fig:2:overview}, bottom-right), each aimed specifically at reducing \texttt{gather} bandwidth.

\noindent\textbf{1) Q/K/V channel compression.} Inspired by multi-head latent attention~\cite{deepseekai2024deepseekv2strongeconomicalefficient}, the query, key, and value channels are compressed by $4\times$ before entering the MatchAttention core, reducing the head dimension from $c/h$ to $c/(4h)$. This reduces gather bandwidth by $4\times$. Under the MatchAttentionRT configuration, this reduces each head to only two channels, yet the accuracy remains competitive (Section~\ref{sec:4.6:hw_acceleration}), indicating that the explicit matching constraint is effective even at severely reduced channel capacity.

\noindent\textbf{2) Unified cross-scale window.} We set the window size to $w \!=\! 3$ at every scale, yielding a $(w+1)^2 \!=\! 4\!\times\!4$ attention window. A unified window size across scales produces regular memory-access patterns that improve coalescing and reduce gather latency.

\noindent\textbf{3) Horizontal matching for stereo.} Rectified stereo confines valid correspondences to the epipolar line, so the vertical extent of the attention sampling window is redundant. We further reduce the window from $4\!\times\!4$ to a purely horizontal $1\!\times\!4$, eliminating vertical overlap and reducing \texttt{gather} bandwidth by another $4\times$. CAS accordingly degenerates into a 1D variant with two sub-windows (left and right). This optimization is stereo-specific and is not applied to optical flow.

\noindent\textbf{Cumulative effect.} Channel compression ($4\times$) and horizontal matching ($4\times$) together yield a $16\times$ reduction in \texttt{gather} bandwidth. After all three optimizations, the share of MatchAttention in the wall-clock latency drops from $>$70\% to $\sim$13\%, approaching the theoretical lower bound. We denote the resulting model as \textbf{MatchAttentionRT}. As shown in Section~\ref{sec:4.6:hw_acceleration}, these optimizations yield a $\sim$3$\times$ end-to-end speedup with minimal accuracy drop, and MatchAttentionRT also surpasses recent non-real-time zero-shot baselines \cite{zerostereo} on multiple datasets.

\revisionon
\section{Experiments}
\label{sec:4:experiments}

We evaluate two instantiations that share the same MatchAttention Block. MatchAttentionXL targets accuracy, and MatchAttentionRT targets real-time edge inference. The two models differ only in channel width and in whether 2D MatchAttention is used. Their architectural details are summarized in Table~\ref{tab:1:arc_detail}. The experiments assess the benchmark accuracy, the zero-shot generalization, the real-time performance on edge devices, and the contribution of each design choice through an ablation study.

\subsection{Experiment Setup}
\subsubsection{Datasets and Metrics}
We evaluate on Middlebury V3 \cite{middlebury} (high-resolution indoor scenes with dense structured-light ground truth), KITTI 2012 \cite{kitti2012} and KITTI 2015 \cite{kitti2015} (real-world driving scenes with sparse LiDAR ground truth), and ETH3D \cite{eth3d} (indoor and outdoor scenes with ground truth obtained using a high-precision laser scanner). For the metrics, ``bad $X.0$'' denotes the percentage of pixels whose disparity error exceeds $X$ pixels; ``EPE'' is the mean absolute disparity error across all evaluated pixels; and ``D1'' is the standard KITTI outlier metric (error $>$\,3\,px and $>$\,5\% of the ground truth). ``Noc'' and ``All'' indicate evaluation on non-occluded and all regions, respectively. We use ``Noc'' as the default unless otherwise noted, since this is the evaluation mask used by most public benchmarks.

\subsubsection{Implementation Details}
MatchAttention is implemented in PyTorch with three backends: a pure-PyTorch fallback, a custom fused CUDA kernel, and \texttt{torch.compile} with FP16 for real-time deployment. The consistency-check threshold $A$ in Eq.~\ref{eq:3:consistency_check} is set to $1$ pixel, and the consistency weight $\epsilon$ in Eq.~\ref{eq:3:loss_cross} is set to $0.01$. A learnable two-dimensional scale $\boldsymbol{\beta}$ balances the numerical ranges of cross-$R_{pos}$ and token features before concatenation, stabilizing gradient propagation.

\begin{table}[]
\revisionon
    \centering
    \caption{Architecture details of MatchAttentionXL and MatchAttentionRT. Encoder scales span $1/4$ to $1/32$; decoder scales span $1/32$ to $1/4$.}
    \label{tab:1:arc_detail}
        \footnotesize
        \setlength{\tabcolsep}{1pt}
        \begin{tabular*}{\columnwidth}{@{\extracolsep{\fill}}lcc}
            \toprule[1pt]
\revisionon
            Architecture & MatchAttentionXL & MatchAttentionRT \\
            \midrule
            Encoder depths            & (2, 2, 6, 2)         & (2, 2, 6, 2) \\
            Encoder channels          & (384, 768, 1024, 1536) & (32, 64, 128, 256) \\
            Decoder depths            & (8, 8, 8, 2)         & (8, 8, 8, 2) \\
            Decoder channels          & (1536, 1024, 768, 384) & (256, 128, 64, 32) \\
            Window size              & $4{\times}4$ & $1{\times}4$ \\
            MLP / ConvGLU ratio       & 2 / 2                & 2 / 2 \\
            Number of heads                 & 4                    & 4 \\
            Channel compression       & $4\times$                   & $4\times$ \\
            \bottomrule[1pt]
        \end{tabular*}
\end{table}

All models are trained from scratch (no pre-trained encoders) with the AdamW optimizer \cite{adamw}, weight decay $0.05$, and a one-cycle learning rate schedule. The training crop is $512\times768$ with the data augmentations of RAFT-Stereo \cite{raftstereo}. Two training-data regimes are used throughout the experiments: (i)~\emph{Scene~Flow~only} trains on Scene Flow \cite{sceneflow} alone for 200k steps with batch size 32; (ii)~\emph{FSD~Mix} trains on FSD \cite{foundationstereo}, CREStereo \cite{cre}, WMGStereo \cite{wmgstereo}, IRS \cite{irs}, Scene Flow \cite{sceneflow}, Sintel \cite{sintel}, and InStereo2K \cite{instereo2k} for 1000k steps with batch size 32. Both regimes are evaluated zero-shot unless fine-tuning is specified. For the ablation study (Section~\ref{sec:4.5:ablation}), MatchAttentionRT is retrained on FSD for 200k steps with batch size 32. Benchmark fine-tuning uses 100k steps on a mixture of FSD Mix and Middlebury, and 100k steps on KITTI 2012, KITTI 2015, and Virtual KITTI 2 \cite{vkitti2}. ETH3D is evaluated zero-shot without fine-tuning. Optical-flow results and the full per-scene Middlebury tables are provided in the supplementary material.

\subsection{Comparison with the State-of-the-Arts}
\label{sec:4.3:benchmark}

MatchAttentionXL is compared with three categories of state-of-the-art methods. Cost-volume iterative methods include RAFT-Stereo~\cite{raftstereo}, IGEV-Stereo~\cite{igev}, and Selective-IGEV~\cite{selective}. Foundation-model-augmented methods include FoundationStereo~\cite{foundationstereo}, AIO-Stereo~\cite{AIO}, and DEFOM-Stereo~\cite{defom_stereo}. Other recent top-ranked Middlebury entries include S2M2~\cite{s2m2}, MGS-Stereo~\cite{mgsstereo}, and StereoAnywhere~\cite{stereo_anywhere}. These methods represent diverse and competitive solutions for high-accuracy stereo. Cost-volume iterative methods generally keep a static correlation representation during inference, whereas foundation-model-augmented methods leverage global attention and strong monocular priors.

Most methods in these categories process each view through a separate cost representation and estimate one-view disparity per forward pass. In contrast, MatchAttention uses the symmetry of epipolar matching to cross-update the features and cross-$R_{pos}$ of both views at once, producing the left and right disparities in a single forward pass, and it embeds the matching constraint directly into the attention operator. A lightweight all-pair correlation is computed only at $1/32$ resolution to initialize cross-$R_{pos}$ with a coarse starting point. For input resolutions above 2K, we instead run the full network at half resolution and use its output, resized to $1/32$, as the initial correlation. Matching is thereafter driven entirely by MatchAttention, without a separately constructed cost volume or external monocular priors.

% Auto-generated by scripts/build_middv3_anchor.py — do not edit by hand.
\begin{table*}[!t]
\revisionon
    \centering
    \caption{Per-scene zero-shot accuracy on the Middlebury~V3 benchmark at full resolution (Noc mask). Best per column in yellow bold, second-best in soft blue. Within each panel, baselines are ordered by weighted-average ascending; MatchAttentionXL is listed in the last row.}
    \label{tab:2:middv3_perscene}
        \footnotesize
        \setlength{\tabcolsep}{1.5pt}
        \begin{tabular*}{\textwidth}{@{\extracolsep{\fill}}lcccccccccccccccc}
            \toprule[1pt]
            \multirow{2}{*}{\textbf{Method}} & \multirow{2}{*}{\textbf{Avg}} & \multicolumn{15}{c}{\textbf{Per-scene} $\downarrow$} \\
            \cmidrule(lr){3-17}
            & & Aust & AustP & Bcyc & Clas & ClaE & Comp & Crus & CruP & Djmb & DjmL & Hoop & Lvgr & Nkub & Plnt & Stair \\
            \midrule
            \multicolumn{17}{l}{\textit{(a) Bad 0.5\,(\%) $\downarrow$}} \\
            \midrule
            S2M2 \cite{s2m2} & \second{22.1} & 23.4 & 17.2 & \best{8.28} & 18.6 & \best{45.0} & 45.2 & 27.0 & 26.1 & 13.7 & 19.4 & 20.1 & 13.0 & \second{30.0} & \best{21.4} & \best{3.92} \\
            FoundationStereo \cite{foundationstereo} & 22.5 & \best{11.1} & 10.7 & 8.66 & 17.9 & 61.5 & \best{36.2} & \second{21.3} & \second{20.7} & \second{7.82} & \second{16.9} & \second{18.2} & 12.1 & 50.7 & 36.6 & \second{10.5} \\
            MGS-Stereo \cite{mgsstereo} & 23.4 & 17.0 & 15.7 & 9.43 & 16.4 & \second{52.1} & 38.4 & 32.6 & 32.7 & 12.4 & 24.0 & 24.5 & 13.2 & \best{29.6} & 26.8 & 11.8 \\
            AIO-Stereo \cite{AIO} & 24.0 & 16.6 & 14.7 & 10.6 & \second{13.0} & 59.8 & 38.5 & 36.5 & 35.2 & 7.92 & 24.8 & 22.4 & 13.7 & 32.2 & 29.4 & 13.6 \\
            Selective-IGEV \cite{selective} & 24.6 & 15.2 & 14.6 & 10.3 & 14.2 & 61.5 & 38.7 & 36.8 & 35.6 & 9.93 & 29.0 & 25.3 & 12.2 & 35.6 & \second{25.6} & 17.6 \\
            DEFOM-Stereo \cite{defom_stereo} & 25.2 & \second{11.6} & 21.4 & 9.65 & 17.8 & 55.0 & 37.3 & 34.6 & 34.1 & 12.1 & 24.6 & 27.0 & \second{11.6} & 32.9 & 38.0 & 13.7 \\
            StereoAnywhere \cite{stereo_anywhere} & 27.3 & 27.2 & \second{7.88} & 12.1 & 17.4 & 58.4 & 45.0 & 37.0 & 32.3 & 8.72 & 28.4 & 30.0 & 13.4 & 50.0 & 35.3 & 20.3 \\
            MatchAttentionXL & \best{21.3} & 15.8 & \best{7.79} & \second{8.61} & \best{10.1} & 55.3 & \second{37.2} & \best{16.5} & \best{18.0} & \best{5.89} & \best{9.76} & \best{13.9} & \best{8.17} & 48.9 & 47.7 & 19.3 \\
            \midrule
            \multicolumn{17}{l}{\textit{(b) Bad 1.0\,(\%) $\downarrow$}} \\
            \midrule
            S2M2 \cite{s2m2} & \second{3.57} & \best{2.81} & \best{2.33} & 3.50 & 2.33 & \best{3.18} & 2.07 & \second{2.58} & \second{2.75} & 1.97 & \second{4.12} & 9.24 & 4.44 & 9.36 & \best{3.18} & \best{0.87} \\
            FoundationStereo \cite{foundationstereo} & 4.39 & 4.23 & 2.74 & \second{3.13} & \second{2.10} & 11.4 & \best{1.57} & 2.83 & 2.86 & 1.84 & 6.34 & \second{6.97} & 5.30 & 11.7 & 4.75 & \second{3.13} \\
            MGS-Stereo \cite{mgsstereo} & 5.69 & 4.89 & 3.14 & 4.85 & 3.28 & 16.3 & 3.54 & 5.83 & 5.72 & \second{1.83} & 6.04 & 11.1 & 4.26 & 12.1 & 5.27 & 4.17 \\
            DEFOM-Stereo \cite{defom_stereo} & 5.81 & 4.57 & 3.92 & 4.79 & 3.35 & 14.3 & 3.04 & 5.67 & 5.50 & 2.41 & 7.54 & 15.3 & \second{3.98} & 11.3 & 5.71 & 4.22 \\
            AIO-Stereo \cite{AIO} & 6.08 & 4.54 & 3.16 & 5.98 & 2.76 & 16.2 & 3.52 & 7.97 & 7.67 & 2.29 & 10.4 & 10.7 & 4.64 & \best{8.39} & 5.34 & 6.76 \\
            Selective-IGEV \cite{selective} & 6.53 & 4.68 & 3.16 & 5.32 & 3.86 & 17.0 & 3.60 & 8.84 & 8.12 & 2.74 & 11.5 & 10.7 & 5.12 & \second{9.05} & 5.86 & 8.00 \\
            StereoAnywhere \cite{stereo_anywhere} & 7.99 & 13.0 & 3.49 & 5.40 & 7.96 & 28.2 & 3.51 & 8.70 & 6.20 & 2.37 & 12.1 & 13.1 & 5.83 & 15.0 & 5.60 & 5.16 \\
            MatchAttentionXL & \best{3.49} & \second{3.20} & \second{2.47} & \best{2.43} & \best{1.61} & \second{6.20} & \second{1.91} & \best{2.43} & \best{2.21} & \best{1.33} & \best{2.60} & \best{4.58} & \best{1.87} & 12.3 & \second{4.70} & 4.26 \\
            \bottomrule[1pt]
        \end{tabular*}
\end{table*}

On the public Middlebury V3 benchmark (Table~\ref{tab:2:middv3_perscene}), MatchAttentionXL ranks among the top three published methods on all six weighted-average error metrics (bad 0.5, bad 1.0, bad 2.0, bad 4.0, EPE, and RMS). The full six-metric breakdown is given in the supplementary material. The accuracy is consistent from the sub-pixel bad 0.5 metric to the coarse bad 4.0 metric, which indicates that the explicit matching constraint does not improve one error metric at the expense of another. The per-scene results show that the advantage is largest at the finest granularity, where MatchAttentionXL obtains the best per-scene result on most of the 15 scenes under bad 0.5 and bad 1.0, where sub-pixel precision is most demanding.

\begin{table*}[]
\revisionon
    \centering
    \caption{Benchmark performance on KITTI 2012, KITTI 2015, ETH3D, and Middlebury. Metrics are reported from the official benchmark websites. Bad 3.0 (3px) and D1 are the default KITTI 2012 and KITTI 2015 metrics, respectively. ETH3D results for MatchAttentionXL are zero-shot (trained on FSD Mix, without ETH3D fine-tuning). ``--'' denotes methods that did not submit to that benchmark.}
    \label{tab:3:benchmark}
        \footnotesize
        \setlength{\tabcolsep}{2pt}
        \begin{tabular*}{\textwidth}{@{\extracolsep{\fill}}lccccccccccc}
            \toprule[1pt]
            \multirow{2}{*}{\textbf{Methods}} & \multicolumn{2}{c}{\textbf{KITTI12}} & \multicolumn{2}{c}{\textbf{KITTI15}} &
            \multicolumn{3}{c}{\textbf{ETH3D}} & \multicolumn{4}{c}{\textbf{Middlebury}} \\
            \cmidrule(lr){2-3} \cmidrule(lr){4-5} \cmidrule(lr){6-8} \cmidrule(l){9-12}
            & 2px $\downarrow$ & 3px $\downarrow$ & D1-bg $\downarrow$ & D1 $\downarrow$ & Bad 1.0 $\downarrow$ & EPE $\downarrow$ & RMS $\downarrow$ & Bad 0.5 $\downarrow$ & Bad 1.0 $\downarrow$ & EPE $\downarrow$ & RMS $\downarrow$ \\
            \midrule
            S2M2 \cite{s2m2} & -- & -- & 2.51 & 3.12 & \best{0.22} & 0.10 & 0.20 & \second{22.1} & \second{3.57} & \second{0.69} & 6.01 \\
            FoundationStereo \cite{foundationstereo} & -- & -- & -- & -- & 0.26 & \second{0.09} & 0.20 & 22.5 & 4.39 & 0.78 & 6.48 \\
            MGS-Stereo \cite{mgsstereo} & -- & -- & -- & -- & 1.10 & 0.14 & 0.32 & 23.4 & 5.69 & 0.74 & 5.91 \\
            AIO-Stereo \cite{AIO} & 1.58 & 1.05 & 1.22 & 1.43 & 0.94 & 0.13 & 0.30 & 24.0 & 6.08 & 0.85 & 6.78 \\
            Selective-IGEV \cite{selective} & 1.59 & 1.07 & 1.22 & 1.44 & 1.23 & 0.12 & 0.29 & 24.6 & 6.53 & 0.91 & 7.26 \\
            DEFOM-Stereo \cite{defom_stereo} & \second{1.43} & \second{0.94} & \second{1.15} & \best{1.33} & 0.70 & 0.11 & 0.22 & 25.2 & 5.81 & 0.79 & \second{5.81} \\
            StereoAnywhere \cite{stereo_anywhere} & -- & -- & -- & -- & -- & -- & -- & 27.3 & 7.99 & 0.93 & 6.18 \\
            \midrule
            MatchAttentionXL & \best{1.31} & \best{0.87} & \best{1.11} & \second{1.40} & \second{0.26} & \best{0.09} & \best{0.18} & \best{21.3} & \best{3.49} & \best{0.68} & \best{5.66} \\
            \bottomrule[1pt]
        \end{tabular*}
\end{table*}

Table~\ref{tab:3:benchmark} extends the comparison to four official leaderboards. MatchAttentionXL attains the best result on the default KITTI~2012 metric and ranks first on all four Middlebury metrics (bad~0.5, bad~1.0, EPE, RMS). On KITTI~2015, it ranks first on D1-bg and second on the default D1, behind DEFOM-Stereo, which also employs a depth foundation model. On ETH3D, MatchAttentionXL is evaluated zero-shot (trained on FSD~Mix, without ETH3D fine-tuning) and obtains the best EPE and RMS, while its bad~1.0 is second to S2M2. Several baselines omit submissions on individual benchmarks (denoted ``--''), but among methods with broad coverage the ranking remains consistent with the Middlebury results in Table~\ref{tab:2:middv3_perscene}. The consistent ranking across benchmarks with different image resolutions, scene types, and ground-truth densities indicates stable cross-benchmark behavior.

\begin{table*}[!t]
\revisionon
    \centering
    \caption{Zero-shot stereo generalization on KITTI 2015, Middlebury, and ETH3D. All methods are trained from scratch on Scene Flow and evaluated zero-shot. All and Noc denote all-pixel and non-occluded evaluation masks, respectively. Best per column in yellow bold, second-best in soft blue.}
    \label{tab:4:zero_shot_stereo}
        \footnotesize
        \setlength{\tabcolsep}{3pt}
        \begin{tabular*}{\textwidth}{@{\extracolsep{\fill}}lcccccccccc}
            \toprule[1pt]
            \multirow{3}*[-3pt]{\textbf{Method}} &
            \multicolumn{2}{c}{\textbf{KITTI15}} & \multicolumn{6}{c}{\textbf{Middlebury}} & \multicolumn{2}{c}{\textbf{ETH3D}} \\
            & \multicolumn{2}{c}{\textbf{Bad 3.0} $\downarrow$}
            & \multicolumn{2}{c}{\textbf{F (Bad 2.0)} $\downarrow$}
            & \multicolumn{2}{c}{\textbf{H (Bad 2.0)} $\downarrow$}
            & \multicolumn{2}{c}{\textbf{Q (Bad 2.0)} $\downarrow$}
            & \multicolumn{2}{c}{\textbf{Bad 1.0} $\downarrow$} \\
            \cmidrule(lr){2-3} \cmidrule(lr){4-5} \cmidrule(lr){6-7} \cmidrule(lr){8-9} \cmidrule(l){10-11}
            & All & Noc & All & Noc & All & Noc & All & Noc & All & Noc \\
            \midrule
            DSMNet \cite{DSM}                       & 5.50 & 5.19 & 41.96 & 38.54 & 18.74 & 14.49 & 13.75 & 9.44 & 4.03 & 3.62 \\
            CFNet \cite{cfnet}                      & 5.99 & 5.79 & 35.21 & 30.05 & 21.99 & 17.69 & 14.21 & 10.51 & 6.08 & 5.48 \\
            Graft-PSMNet \cite{graftnet}            & 5.34 & 5.00 & 39.92 & 36.30 & 17.65 & 13.36 & 13.92 & 9.23 & 11.43 & 10.7 \\
            ITSA-CFNet \cite{itsa}                  & \best{4.73} & \second{4.67} & 34.01 & 30.14 & 16.48 & 12.32 & 12.28 & 8.54 & 5.43 & 5.17 \\
            HVT-PSMNet \cite{hvt}                   & \second{4.84} & \best{4.63} & 40.74 & 37.60 & 15.66 & 12.55 & 10.12 & 7.00 & 6.07 & 5.65 \\
            RAFT-Stereo \cite{raftstereo}           & 5.47 & 5.27 & 15.63 & 11.94 & 11.20 & 8.66 & 10.25 & 7.44 & \second{2.60} & \second{2.29} \\
            IGEV-Stereo \cite{igev}                 & 6.03 & 5.76 & 30.94 & 28.98 & 11.90 & 9.45 & 8.88 & 6.20 & 4.04 & 3.60 \\
            IGEV++ \cite{igev++}                    & 6.24 & 6.02 & 16.19 & 12.71 & 10.58 & 7.76 & 8.93 & 6.27 & 4.66 & 4.14 \\
            NMRF-Stereo \cite{nmrf}                 & 5.31 & 5.14 & 37.63 & 35.25 & 13.36 & 10.9 & \second{7.87} & 5.07 & 3.80 & 3.48 \\
            Mocha-Stereo \cite{mocha}               & 5.97 & 5.73 & 30.23 & 28.26 & \second{10.18} & 9.45 & 7.96 & \second{4.87} & 4.02 & 3.47 \\
            \midrule
            MatchAttentionRT                        & 5.52 & 5.31 & 15.65 & \second{11.29} & 12.71 & \second{8.01} & 12.42 & 8.23 & 5.06 & 4.64 \\
            MatchAttentionXL                       & 5.28 & 5.11 & \best{10.80} & \best{7.33}  & \best{7.70}  & \best{4.71}  & \best{6.94}  & \best{4.64}  & \best{2.14} & \best{1.93} \\
            \bottomrule[1pt]
        \end{tabular*}
\end{table*}

To remove the effect of training-data scale, Table~\ref{tab:4:zero_shot_stereo} trains all methods only on the small synthetic Scene Flow dataset and evaluates them zero-shot on KITTI 2015, Middlebury, and ETH3D. FoundationStereo and DEFOM-Stereo are excluded here because their DepthAnything V2 encoders are pre-trained on orders of magnitude more data than Scene Flow, so a Scene-Flow-only comparison would not be fair. Among the methods trained under the same data constraint, MatchAttentionXL obtains the best result on all columns of Middlebury F, H, Q and ETH3D, and is competitive with the best baseline on KITTI 2015. This controlled comparison removes the confound of large-scale pre-training and shows that the advantage does not depend solely on training-data scale.

\subsection{Real-Time Zero-Shot Comparison}
\label{sec:4.4:zero_shot}

Table~\ref{tab:5:realtime_zero_shot} compares MatchAttentionRT with published real-time stereo methods under two training regimes, namely million-scale data (panel~a) and Scene Flow only (panel~b). The compared methods cover representative efficient-stereo designs, including cost-volume-based iterative methods (Fast-FSD~\cite{fastfoundationstereo}, RT-IGEV++~\cite{igev++}, RT-Monster++~\cite{monster++}) and lightweight feed-forward or implicit-aggregation models (LightStereo~\cite{lightstereo}, LiteAnyStereo~\cite{liteanyStereo}, BANet~\cite{banet}, IINet~\cite{iinet}). Non-real-time methods (Zero-RAFT-Stereo, MonSter++, MatchAttentionXL) are also listed as accuracy references. Accuracy is reported at each dataset's native resolution. Latency is measured at $1024{\times}512$ in FP16 on one desktop GPU and two edge devices, using \texttt{torch.compile} on each method's publicly released code and checkpoint. The suffixes ``-192'' and ``-416'' denote the maximum disparity at inference for methods with an adjustable test-time search range. In terms of implementation, methods that collapse disparity into 2D convolution channels (LiteAnyStereo, LightStereo, BANet-2D, IINet) are fixed to the training-time range, whereas methods using 3D convolutional volumes (Fast-FSD, RT-IGEV++, RT-Monster++, BANet-3D) can vary the range at inference with proportional cost growth. Unlike methods with a fixed cost-volume disparity dimension, the effective disparity range of MatchAttentionRT is governed by the dynamically updated cross-$R_{pos}$, subject to image bounds and convergence.

As shown in Table~\ref{tab:5:realtime_zero_shot}, MatchAttentionRT is the fastest method on the NX-16 among those compared while remaining competitive across all metrics, including Middlebury-F and Middlebury-H (bad 0.5, bad 1.0, bad 2.0), ETH3D (EPE, bad 1.0), and KITTI 2015 (EPE, D1). The most accurate competitor, Fast-FSD-416, is $17\times$ slower on the NX-16, whereas the methods with comparable latency, such as LiteAnyStereo and BANet-2D, are less accurate on most metrics. MatchAttentionRT also surpasses the non-real-time Zero-RAFT-Stereo~\cite{zerostereo} on most metrics and exceeds MonSter++~\cite{monster++} on Middlebury-F while using $46\times$ fewer parameters than MatchAttentionXL, which indicates that real-time inference is achieved here without sacrificing zero-shot generalization. A qualitative comparison on high-resolution Middlebury-F scenes is provided in Fig.~S1 of the supplementary material.

\begin{table*}[!t]
\revisionon
    \centering
    \caption{Real-time zero-shot stereo generalization. Accuracy is evaluated at each benchmark's native resolution; latency is measured at $1024{\times}512$ in FP16 with \texttt{torch.compile} on all devices. NX-16 = Jetson Orin NX 16\,GB; AGX-64 = Jetson Orin AGX 64\,GB. Method suffixes ``-192'' and ``-416'' denote the maximum disparity at inference. Best per column in yellow bold, second-best in soft blue within each subtable.}
    \label{tab:5:realtime_zero_shot}
        \footnotesize
        \setlength{\tabcolsep}{1.8pt}
        \ifdefined\SINGLECOLUMN
          \resizebox{\textwidth}{!}{%
          \begin{tabular}{@{}lcccccccccccccccc@{}}
        \else
          \begin{tabular*}{\textwidth}{@{\extracolsep{\fill}}lcccccccccccccccc}
        \fi
            \toprule[1pt]
            \multirow{3}{*}{\textbf{Method}}
            & \multicolumn{3}{c}{\textbf{Middlebury-F} $\downarrow$}
            & \multicolumn{3}{c}{\textbf{Middlebury-H} $\downarrow$}
            & \multicolumn{2}{c}{\textbf{ETH3D} $\downarrow$}
            & \multicolumn{2}{c}{\textbf{KITTI15} $\downarrow$}
            & \multirow{3}{*}{\shortstack{\textbf{FLOPs}\\(G)}}
            & \multicolumn{3}{c}{\textbf{Latency (ms)} $\downarrow$}
            & \multirow{3}{*}{\shortstack{\textbf{Params}\\(M)}} \\
            \cmidrule(lr){2-4} \cmidrule(lr){5-7} \cmidrule(lr){8-9} \cmidrule(lr){10-11} \cmidrule(lr){13-15}
            & Bad 0.5 & Bad 1.0 & Bad 2.0 & Bad 0.5 & Bad 1.0 & Bad 2.0 & EPE & Bad 1.0 & EPE & D1
            & & \textbf{NX-16} & \textbf{AGX-64} & \textbf{4060Ti} & \\
            \midrule
            \multicolumn{16}{l}{\textit{(a) Trained on million-scale data (each method's largest reported dataset).}} \\
            \midrule
            Fast-FSD-192 \cite{fastfoundationstereo}            & 42.82 & 30.52 & 24.61 & \second{13.77} & \second{7.35} & 4.79          & \best{0.195} & \best{1.21} & \second{0.900} & \second{3.23} & 1145.0          & 777.5            & 299.1          & 222.2          & 17.65 \\
            Fast-FSD-416 \cite{fastfoundationstereo}            & \second{25.30} & \second{11.97} & \second{7.20}  & \best{11.80} & \best{4.81} & \best{2.20}    & \second{0.196} & \second{1.22} & \best{0.896} & \best{3.17}    & 1811.6          & 1315.0             & 497.7          & 421.0          & 17.65 \\
            RT-Monster++-192 \cite{monster++}       & 51.04  & 34.28 & 26.03 & 27.10 & 14.24 & 7.43        & 0.212          & 1.77          & 0.918 & 3.29          & 578.9   & 294.8   & 118.2  & 44.8  & 42.82 \\
            RT-Monster++-416 \cite{monster++}       & 42.49 & 22.94 & 13.61 & 26.34 & 13.10 & 6.16        & 0.212          & 1.77          & 0.918 & 3.29          &  579.1   &  295.0  & 121.2  & 44.8  & 42.82 \\
            LiteAnyStereo \cite{liteanyStereo}       & 58.39 & 40.46 & 30.77 & 36.00 & 19.02 & 10.39        & 0.337          & 3.73          & 0.965 & 3.71          & \best{61.1}    & \second{133.7}   & \second{60.9}  & \second{16.6}  & \best{7.60} \\
            \midrule
            \textbf{MatchAttentionRT}   & \best{25.12} & \best{9.90} & \best{4.38} & 17.10 & 7.55 & \second{3.76} & 0.210 & 1.82 & 0.979 & 4.32 & \second{62.5}  & \best{79.1}      & \best{36.7}    & \best{9.3}     & \second{10.92} \\
            \midrule
            Zero-RAFT-Stereo \cite{zerostereo}       & 28.17  & 13.91 & 8.43 & 16.74 & 8.32 & 4.45        & 0.230          & 2.13          & 0.997 & 4.17          & 6384.5  & 1007.6   &  389.2 & 216.1  & 11.12 \\
            Monster++-416 \cite{monster++}       & 31.56 & 14.06 & 7.36 & 16.06 & 6.22 & 2.13        & 0.160          & 0.88          & 0.880 & 3.23          & 10843.0    & 3581.7   & 1493.6  & 567.1  & 388.7 \\
            MatchAttentionXL       & 17.36 & 5.19 & 2.08 & 9.40 & 3.71 & 1.89        & 0.146          & 0.81          & 0.870 & 3.30          &  5240.2   & 1287.8   & 493.1  & 248.8  & 507.1 \\
            \midrule
            \multicolumn{16}{l}{\textit{(b) Trained on Scene Flow only.}} \\
            \midrule
            IINet \cite{iinet}                   & 63.04 & 46.95 & 36.38 & 41.31 & 25.64 & 16.37        & 4.890          & 22.21         & 1.369          & 7.00          & 185.4           & 101.5            & 44.5           & 13.7           & 19.56 \\
            LightStereo-L \cite{lightstereo}           & 79.09 & 64.57 & 51.49 & 56.33 & 37.27 & 23.89        & 29.355         & 45.84         & 2.808          & 12.09         & 168.4           & 154.3            & 71.1           & 20.4           & 24.29 \\
            RT-IGEV-192 \cite{igev++}           & 51.94 & 38.87 & 32.37 & \second{26.49} & 16.95 & 11.52 & 1.192 & \second{5.67} & 1.359 & \second{5.86} & 549.0 & 346.7 & 143.3 & 50.4 & \second{4.17} \\
            RT-IGEV-416 \cite{igev++}           & \second{41.00} & \second{25.63} & \second{18.27} & \best{25.60} & \second{15.57} & \second{9.76} & 2.374 & 6.16 & 1.768 & 5.89 & 573.9 & 927.1 & 335.4 & 117.1 & 4.17 \\
            BANet-2D \cite{banet}                & 83.87 & 72.02 & 58.54 & 63.15 & 43.78 & 28.46        & 35.943         & 44.88         & 3.806          & 16.06         & 65.1            & \second{76.6}    & \second{47.8}  & \best{8.0}     & 5.46 \\
            BANet-3D-192 \cite{banet}                & 82.31 & 70.31 & 58.34 & 63.30 & 44.91 & 30.12        & 14.130         & 29.35         & 3.278          & 16.59         & 116.0           & 150.1            & 66.8           & 35.4           & \best{3.63} \\
            BANet-3D-416 \cite{banet}              & 77.37 & 62.79 & 48.96 & 64.17 & 46.16 & 31.42        & 52.843         & 39.41         & 9.535          & 20.74         &   171.4         &    248.7         &   108.5         &   71.5         & 3.63 \\
            LiteAnyStereo \cite{liteanyStereo}       & 64.16 & 47.04 & 35.77 & 45.63 & 27.27 & 15.75        & \second{0.764} & 14.42         & \second{1.235} & 5.95          & \best{61.1}     & 133.7            & 60.9           & 16.6           & 7.60 \\
            \midrule
            \textbf{MatchAttentionRT}   & \best{36.64} & \best{19.41} & \best{11.29} & 27.04 & \best{14.29} & \best{8.01} & \best{0.382} & \best{4.64} & \best{1.119} & \best{5.15} & \second{62.5} & \best{79.1} & \best{36.7} & \second{9.3} & 10.92 \\
            \bottomrule[1pt]
        \ifdefined\SINGLECOLUMN
          \end{tabular}%
          }
        \else
          \end{tabular*}
        \fi
\end{table*}

This advantage is robust across training-data scales. Under million-scale training (panel~a), MatchAttentionRT obtains the best Middlebury-F results and is among the fastest methods on the NX-16. Under Scene-Flow-only training (panel~b), where all methods use the same small synthetic dataset, MatchAttentionRT leads on Middlebury-F (bad 0.5, bad 1.0, bad 2.0), Middlebury-H (bad 1.0, bad 2.0), ETH3D (EPE, bad 1.0), and KITTI 2015 (EPE, D1). On ETH3D EPE it roughly halves the error of the next-best real-time baseline. The accuracy is therefore consistent across both data regimes, which suggests that the advantage stems primarily from the MatchAttention operator itself, as examined in the following ablation study.

\subsection{Ablation Study}
\label{sec:4.5:ablation}

To study how the MatchAttention operator and the model capacity each contribute to accuracy, we retrain MatchAttentionRT variants under controlled modifications, summarized in Table~\ref{tab:6:ablation}. Block~(a) (groups i--v) varies the operator design while keeping the parameter count approximately fixed; block~(b) (groups vi--viii) increases parameters or compute to establish an upper-bound reference. The default configuration (top row, $\star$) is MatchAttentionRT.

\definecolor{betterclr}{HTML}{0B6623}   % dark green: method improved
\definecolor{worseclr}{HTML}{B22222}    % firebrick red: method regressed
% Quality-direction markers w.r.t. the default row.
% All metrics here are lower-is-better, so:
%   green up-triangle = value lower than default = method improved
%   red down-triangle = value higher than default = method regressed
\newcommand{\goodmark}{\,{\scriptsize\color{betterclr}$\blacktriangle$}}
\newcommand{\badmark}{\,{\scriptsize\color{worseclr}$\blacktriangledown$}}
\newcommand{\nomark}{\phantom{\badmark}}
\newcommand{\val}[2]{\makebox[2.9em][r]{#1}\makebox[0.85em][l]{#2}}
\newcommand{\bval}[1]{\makebox[2.9em][r]{\textbf{#1}}\makebox[0.85em][l]{\nomark}}
\newcommand{\slashsep}{\hspace{0.6em}}
\newcommand{\triple}[6]{\val{#1}{#2}\slashsep\val{#3}{#4}\slashsep\val{#5}{#6}}
\newcommand{\pair}[4]{\val{#1}{#2}\slashsep\val{#3}{#4}}
\newcommand{\btriple}[3]{\bval{#1}\slashsep\bval{#2}\slashsep\bval{#3}}
\newcommand{\bpair}[2]{\bval{#1}\slashsep\bval{#2}}

\begin{table*}[!t]
\revisionon
    \centering
    \caption{Ablation study of operator design and model capacity for MatchAttentionRT. All metrics are lower-is-better. Markers compare against the default: {\color{betterclr}$\blacktriangle$} improves, {\color{worseclr}$\blacktriangledown$} regresses; bare numeric values match the default after rounding, and dashes denote unavailable results. $^\dagger$ Swin attention exceeds GPU memory on Middlebury-F.}
    \label{tab:6:ablation}
    \footnotesize
    \renewcommand{\arraystretch}{1.12}
    \setlength{\tabcolsep}{4pt}
    \resizebox{\textwidth}{!}{%
    \begin{tabular}{@{}llccccc@{}}
        \toprule[1pt]
        \textbf{Experiment} & \textbf{Variant}
        & \shortstack{\textbf{Middlebury-F}\\Bad 0.5 / 1.0 / 2.0 $\downarrow$}
        & \shortstack{\textbf{Middlebury-H}\\Bad 0.5 / 1.0 / 2.0 $\downarrow$}
        & \shortstack{\textbf{ETH3D}\\EPE / Bad 1.0 $\downarrow$}
        & \shortstack{\textbf{KITTI15}\\EPE / D1 $\downarrow$}
        & \textbf{Params} \\
        \midrule
        \textbf{MatchAttentionRT}\,$\star$ & \emph{(default)}
        & \btriple{29.02}{12.68}{6.19}
        & \btriple{20.75}{9.90}{4.76}
        & \bpair{0.25}{2.61}
        & \bpair{1.05}{4.40}
        & \textbf{10.92M} \\
        \specialrule{0.7pt}{1pt}{1pt}

        \multicolumn{7}{@{}l}{\textit{(a) Operator-design ablations: each row modifies one design choice of the default.}} \\
        \midrule
        \multirow{4}{*}{(i) Attention family}
        & Local attention
        & \triple{60.80}{\badmark}{39.86}{\badmark}{22.02}{\badmark}
        & \triple{41.86}{\badmark}{24.31}{\badmark}{12.51}{\badmark}
        & \pair{3.04}{\badmark}{4.29}{\badmark}
        & \pair{1.08}{\badmark}{5.07}{\badmark}
        & 10.86M \\
        & NAT \cite{hassani2023neighborhood}
        & \triple{61.72}{\badmark}{40.46}{\badmark}{22.82}{\badmark}
        & \triple{41.74}{\badmark}{24.66}{\badmark}{12.78}{\badmark}
        & \pair{3.16}{\badmark}{4.65}{\badmark}
        & \pair{1.09}{\badmark}{5.16}{\badmark}
        & 10.86M \\
        & Swin attention \cite{Liu_2021_ICCV}$^\dagger$
        & \triple{--}{\nomark}{--}{\nomark}{--}{\nomark}
        & \triple{61.27}{\badmark}{40.21}{\badmark}{21.42}{\badmark}
        & \pair{4.32}{\badmark}{7.47}{\badmark}
        & \pair{1.40}{\badmark}{7.08}{\badmark}
        & 10.86M \\
        & Global 1D attention
        & \triple{70.91}{\badmark}{50.93}{\badmark}{31.49}{\badmark}
        & \triple{52.50}{\badmark}{31.18}{\badmark}{16.29}{\badmark}
        & \pair{4.20}{\badmark}{7.08}{\badmark}
        & \pair{1.28}{\badmark}{6.89}{\badmark}
        & 10.86M \\

        \cmidrule(l){1-7}
        \multirow{2}{*}{(ii) MATBlock layout}
        & self-MAT, no cross-$R_{pos}$
        & \triple{30.80}{\badmark}{13.84}{\badmark}{6.80}{\badmark}
        & \triple{23.04}{\badmark}{11.03}{\badmark}{5.32}{\badmark}
        & \pair{0.26}{\badmark}{3.06}{\badmark}
        & \pair{0.99}{\goodmark}{4.37}{\goodmark}
        & 10.91M \\
        & double cross-MAT
        & \triple{29.41}{\badmark}{13.32}{\badmark}{6.87}{\badmark}
        & \triple{20.83}{\badmark}{10.26}{\badmark}{5.24}{\badmark}
        & \pair{0.25}{\nomark}{2.56}{\goodmark}
        & \pair{0.99}{\goodmark}{4.34}{\goodmark}
        & 11.13M \\

        \cmidrule(l){1-7}
        (iii) Similarity & dot product
        & \triple{29.52}{\badmark}{12.84}{\badmark}{6.14}{\goodmark}
        & \triple{21.22}{\badmark}{10.10}{\badmark}{5.01}{\badmark}
        & \pair{0.25}{\nomark}{2.72}{\badmark}
        & \pair{1.00}{\goodmark}{4.36}{\goodmark}
        & 10.92M \\
        (iv) Window size & $1 \times 8$
        & \triple{31.41}{\badmark}{13.94}{\badmark}{6.84}{\badmark}
        & \triple{22.83}{\badmark}{10.71}{\badmark}{5.08}{\badmark}
        & \pair{0.28}{\badmark}{3.06}{\badmark}
        & \pair{1.09}{\badmark}{4.58}{\badmark}
        & 10.98M \\
        (v) Occlusion & no occlusion handling
        & \triple{31.03}{\badmark}{14.06}{\badmark}{6.96}{\badmark}
        & \triple{24.29}{\badmark}{11.49}{\badmark}{5.32}{\badmark}
        & \pair{0.26}{\badmark}{3.08}{\badmark}
        & \pair{0.99}{\goodmark}{4.28}{\goodmark}
        & 10.75M \\

        \specialrule{0.7pt}{2pt}{1pt}
        \multicolumn{7}{@{}l}{\textit{(b) Capacity-scaling ablations: each row increases parameters or compute.}} \\
        \midrule
        (vi) Width & 2$\times$ channel widths
        & \triple{26.63}{\goodmark}{11.13}{\goodmark}{5.38}{\goodmark}
        & \triple{18.07}{\goodmark}{8.20}{\goodmark}{3.77}{\goodmark}
        & \pair{0.22}{\goodmark}{2.09}{\goodmark}
        & \pair{0.99}{\goodmark}{4.25}{\goodmark}
        & 42.61M \\
        (vii) 1D vs 2D & 2D MAT
        & \triple{28.45}{\goodmark}{12.26}{\goodmark}{5.80}{\goodmark}
        & \triple{20.68}{\goodmark}{9.74}{\goodmark}{4.67}{\goodmark}
        & \pair{0.24}{\goodmark}{2.40}{\goodmark}
        & \pair{1.05}{\nomark}{4.42}{\badmark}
        & 11.10M \\
        \multirow{2}{*}{\begin{tabular}{@{}c@{\hspace{0.4em}}l@{}}\multirow{2}{*}{(viii)} & Channel\\ & compression\end{tabular}}
        & full channels (1D MAT)
        & \triple{28.19}{\goodmark}{12.17}{\goodmark}{5.89}{\goodmark}
        & \triple{20.69}{\goodmark}{9.56}{\goodmark}{4.41}{\goodmark}
        & \pair{0.23}{\goodmark}{2.34}{\goodmark}
        & \pair{1.00}{\goodmark}{4.40}{\nomark}
        & 16.50M \\
        & full channels + 2D MAT
        & \triple{25.95}{\goodmark}{10.87}{\goodmark}{4.96}{\goodmark}
        & \triple{18.05}{\goodmark}{8.02}{\goodmark}{3.71}{\goodmark}
        & \pair{0.22}{\goodmark}{1.85}{\goodmark}
        & \pair{1.03}{\goodmark}{4.23}{\goodmark}
        & 16.68M \\

        \bottomrule[1pt]
    \end{tabular}}
\end{table*}

Replacing MatchAttention with another attention family consistently degrades the reported metrics by a large margin (Table~\ref{tab:6:ablation}, group~i). Local attention and NAT~\cite{hassani2023neighborhood} lack long-range connections and increase the Middlebury-F bad 1.0 error by approximately 27--28 percentage points. Global 1D attention covers the full disparity range but loses the explicit matching constraint, raising Middlebury-F bad 1.0 by approximately 38 percentage points. Swin attention~\cite{Liu_2021_ICCV} exceeds the GPU memory on Middlebury-F. It should be noted that methods based on global 1D attention (S2M2~\cite{s2m2}) or Swin attention (UniMatch~\cite{xu2023unifying}) require an explicit cost volume and cost-volume-level supervision to compensate, instead of directly regressing cross-$R_{pos}$ from attention as in our approach. This comparison across attention families shows that explicit, dynamic, long-range matching cannot be obtained from these prior attention designs alone.

The effect of the CAS window size is tested by doubling the window from $1\times4$ to $1\times8$ (group~iv). Although the larger window covers a strictly larger receptive field, every metric degrades on Middlebury, ETH3D, and KITTI 2015, which is consistent with the ``match is always inside the window'' failure mode discussed in Fig.~\ref{fig:1:attention_comparison} and confirms that a smaller window produces a stronger explicit-matching signal. The remaining design choices (groups~ii, iii, v) each address a distinct error mode. Removing cross-$R_{pos}$ from the self-MAT input (group~ii) increases the Middlebury-H bad 1.0 error by approximately 11\% by eliminating the label-propagation effect (Section~\ref{sec:3.2:block}). Replacing the negative-$L_1$ similarity with a dot product (group~iii) gives small gains on a few rounded metrics but regresses most Middlebury-H and ETH3D metrics, indicating that the negative-$L_1$ similarity is the more stable choice across datasets. Disabling occlusion handling (group~v) degrades the results on Middlebury and ETH3D, whose indoor scenes contain severe occlusions, but has a negligible effect on KITTI 2015, where outdoor driving scenes have sparse occlusion boundaries and the consistency-constrained loss provides little additional signal. These results show that each component of the MATBlock contributes to the final accuracy.

\begin{table*}[!t]
\revisionon
    \centering
    \caption{Accuracy and latency of four acceleration configurations at $1024{\times}512$ (FP16, \texttt{torch.compile}). The starred configuration ($1/4$ channels and 1D MatchAttention) is MatchAttentionRT, the only one that runs in real time on Jetson Orin NX 16\,GB. NX-16 = Orin NX 16\,GB; AGX-64 = Orin AGX 64\,GB; 4060Ti = RTX 4060 Ti. Best per column in yellow bold, second-best in soft blue.}
    \label{tab:7:acceleration_anchor}
        \footnotesize
        \setlength{\tabcolsep}{6pt}
        \begin{tabular*}{0.95\textwidth}{@{\extracolsep{\fill}}lccccccc}
            \toprule[1pt]
            \multirow{2}{*}{\textbf{Configuration}}
            & \textbf{Middlebury-H} & \textbf{ETH3D}
            & \multirow{2}{*}{\shortstack{\textbf{FLOPs}\\(G)\,$\downarrow$}}
            & \multicolumn{3}{c}{\textbf{Latency (ms)} $\downarrow$}
            & \multirow{2}{*}{\textbf{Params}\,$\downarrow$} \\
            \cmidrule(lr){5-7}
            & Bad 2.0\,$\downarrow$ & Bad 1.0\,$\downarrow$
            & & \textbf{NX-16} & \textbf{AGX-64} & \textbf{4060Ti} & \\
            \midrule
            Full channels + 2D MAT          & \best{3.71} & \best{1.85} & 92.6          & 213.9         & 85.6          & 25.0          & 16.68M \\
            Full channels + 1D MAT          & \second{4.41} & \second{2.34} & 90.7        & \second{97.4} & \second{43.2} & \second{11.4} & 16.50M \\
            $1/4$ channels + 2D MAT         & 4.67          & 2.40          & \second{64.4} & 143.2       & 62.1          & 17.4          & \second{11.10M} \\
            $1/4$ channels + 1D MAT $\star$ & 4.76          & 2.61          & \best{62.5} & \best{79.1}   & \best{36.7}   & \best{9.3}    & \best{10.92M} \\
            \bottomrule[1pt]
        \end{tabular*}
\end{table*}

\begin{figure}[!t]
\revisionon
    \centering
    \includegraphics[width=\columnwidth]{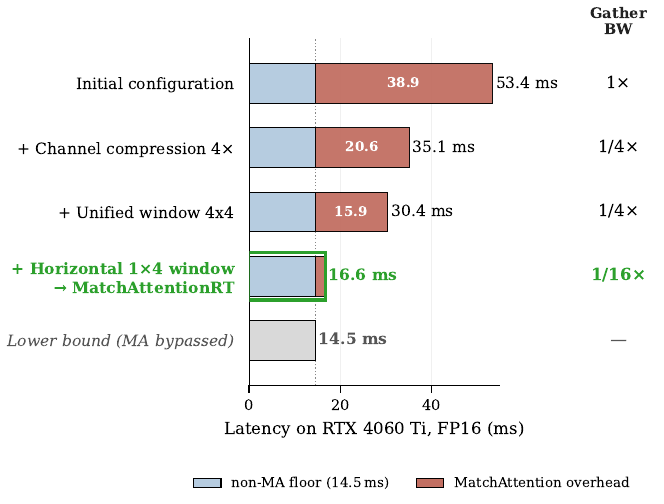}
    \caption{Hardware-aware acceleration cascade on RTX~4060 Ti at $1280{\times}736$ (FP16, \texttt{torch.compile}). The initial unaccelerated 2D configuration (top) uses window size $[6{\times}6, 6{\times}6, 4{\times}4, 4{\times}4]$ and no channel compression. Each step adds one optimization, ending at MatchAttentionRT.}
    \label{fig:5:acceleration_cascade}
\end{figure}

Block~(b) (groups vi--viii) shows that capacity scaling improves most metrics by modest absolute margins, with Middlebury bad-rate gains mostly around 1--3 percentage points even for the strongest configuration (full channels + 2D MatchAttention). The $2\times$-width configuration (group~vi, 42.61M parameters) underperforms the full-channels + 2D configuration (group~viii, 16.68M parameters) on Middlebury and ETH3D despite using $2.6\times$ more parameters. Together with the operator ablations in block~(a), these results indicate that accuracy is determined primarily by the MatchAttention operator rather than by model capacity.

\subsection{Efficiency Analysis and Edge Deployment}
\label{sec:4.6:hw_acceleration}

The low latency of MatchAttentionRT is achieved by three hardware-aware optimizations (Section~\ref{sec:3.4:acceleration}). As shown in Fig.~\ref{fig:5:acceleration_cascade}, although MatchAttention is linear in complexity, it has a practical bottleneck. The non-parameterized modules account for only 3 to 5\% of the FLOPs, but their per-query dynamic indexing of cross-$R_{pos}$ and self-$R_{pos}$ produces random-access \texttt{gather} operations that dominate the wall-clock time. Channel compression reduces the \texttt{gather} bandwidth by $4\times$, a unified $4\times4$ window aligns memory access across scales, and the stereo-specific horizontal $1\times4$ window brings the cumulative reduction to $16\times$, lowering the wall-clock share of MatchAttention from over 70\% to approximately 13\%.

Table~\ref{tab:7:acceleration_anchor} compares four configurations formed by \{full, $1/4$ channels\} and \{1D, 2D MatchAttention\} in terms of accuracy and efficiency. The accuracy spread is small (approximately 1 percentage point on Middlebury-H bad 2.0), while latency varies substantially across devices. We adopt the $1/4$-channel + 1D configuration as MatchAttentionRT because it is the fastest on the NX-16 while remaining within 1 to 3 percentage points of the strongest configuration on all metrics (Section~\ref{sec:4.5:ablation}). To the best of our knowledge, MatchAttentionRT is the first stereo model that runs in real time on a 16\,GB embedded device while retaining zero-shot generalization competitive with non-real-time state-of-the-art methods.

\revisionon
\section{Conclusion}
\label{sec:5:conclusion}

This paper has introduced MatchAttention, a correspondence-aware attention mechanism that embeds an explicit matching constraint into stereo matching. The key idea is to treat the relative position between a query and its matched key as a learnable state inside the attention operator, rather than as an auxiliary positional embedding or a post-attention prediction target. Cross-$R_{pos}$ centers a compact CAS window on the current correspondence estimate, while self-$R_{pos}$ propagates correspondence cues within each view. This design decouples matching range from window size, enabling long-range, discriminative, sub-pixel correspondence refinement at strictly linear complexity.

Built on this mechanism, the hierarchical MatchAttention network provides a unified path to accuracy and efficiency. The high-accuracy variant achieves leading performance across standard stereo benchmarks, while the real-time variant runs on embedded GPUs and retains strong zero-shot generalization. Controlled training and ablation results show that these gains are not primarily due to dataset scale, external pretraining, or model capacity, but to the explicit, dynamic, long-range matching constraint built directly into attention.

The supplementary optical flow results indicate that MatchAttention extends beyond rectified stereo. Extending the formulation to multi-view reconstruction and temporal correspondence is a promising direction. Future work may further improve the accuracy--efficiency frontier through model compression such as pruning, distillation, or architecture search.

% === Appendix (moved to supplementary material) ===

\iffalse

\fi

% % use section* for acknowledgment
% \ifCLASSOPTIONcompsoc
%   % The Computer Society usually uses the plural form
%   \section*{Acknowledgments}
% \else
%   % regular IEEE prefers the singular form
% \fi

\ifCLASSOPTIONcaptionsoff
  \newpage
\fi

\bibliographystyle{IEEEtran}
\bibliography{IEEEabrv,mybibfile}
%% === Supplementary Material (compiled into the same PDF) ===
\revisionon
% Renumber sections / subsections / tables / figures / equations with an
% S-prefix so the supplementary content has its own numbering line.
\clearpage
\setcounter{section}{0}
\setcounter{subsection}{0}
\setcounter{table}{0}
\setcounter{figure}{0}
\setcounter{equation}{0}
\renewcommand{\thesection}{S\arabic{section}}
\renewcommand{\thesubsection}{\thesection.\arabic{subsection}}
\renewcommand{\thetable}{S\arabic{table}}
\renewcommand{\thefigure}{S\arabic{figure}}
\renewcommand{\theequation}{S\arabic{equation}}

\ifdefined\SINGLECOLUMN
  \section*{Supplementary Material}
\else
  \twocolumn[
   \begin{center}
     {\Large\bfseries Supplementary Material\par}
     \vspace{1em}
   \end{center}
  ]
\fi

\section{Initial Correlation}
\label{sec:s1:init_corr}

The initial cross-$R_{pos}$ is estimated at $1/32$ resolution through a lightweight correlation module shared by stereo matching and optical flow. For stereo, encoder features are matched along the epipolar line; for optical flow, they are matched through a 4D all-pair correlation over spatial positions in both frames.

Before correlation, a 2-layer global attention block, consisting of self-attention, cross-attention, and an FFN in each layer, aggregates context across both views at $1/32$ resolution. The correlation is then converted by a Sinkhorn optimal transport layer into a doubly-stochastic transport matrix, from which a local-window estimator regresses the initial cross-$R_{pos}$. In stereo, the estimated $x$-displacement defines the initial disparity and the $y$-component is set to zero; in optical flow, the 2D displacement forms the initial flow vector directly. The module is supervised at $1/32$ during training and adds negligible runtime overhead compared with the full MatchAttention hierarchy.

\revisionon
\section{MatchAttention for Optical Flow}
\label{sec:s2:flow}

Although the main paper focuses on stereo matching, MatchAttention is not specific to epipolar geometry. Cross-$R_{pos}$ is simply a 2D vector whose two components directly form the optical flow vector when applied to temporally adjacent frames. We describe the minimal changes that adapt MatchAttention to optical flow and report zero-shot results on Sintel, KITTI~2015, and the high-resolution Spring dataset.

\subsection{Model Configuration}
\label{sec:s2.1:arch}

The optical flow model uses MatchAttentionXL with all-pair correlation at $1/32$ in place of the epipolar-line correlation used for stereo, and all other architectural components stay unchanged.

\subsection{Training Protocol}
\label{sec:s2.2:training}

For zero-shot evaluation on Sintel \cite{sintel} and KITTI 2015 \cite{kitti2015}, MatchAttention is trained on Scene Flow \cite{sceneflow} for 100k steps with batch size 32, i.e., the same data used for the stereo zero-shot evaluation in the main paper. We intentionally do not follow the standard C+T protocol (FlyingChairs \cite{flyingchairs} + FlyingThings3D \cite{sceneflow}) because the resolution of FlyingChairs ($384\!\times\!512$) is smaller than our training crop ($512\!\times\!768$). For zero-shot evaluation on the high-resolution Spring dataset \cite{spring}, MatchAttention is further fine-tuned on a mixture of Scene Flow, Sintel, KITTI 2015, and HD1K \cite{hd1k} for 100k steps.

\subsection{Zero-Shot Optical Flow Results}
\label{sec:s2.3:results}

\begin{table}[]
\revisionon
    \centering
    \caption{Zero-shot optical-flow generalization. Sintel results are evaluated with the EPE metric.}
    \label{tab:s1:zero_shot_flow}
        \begin{tabular*}{0.95\columnwidth}{@{\extracolsep{\fill}}lcccc}
            \toprule[1pt]
            \multirow{2}*[-2pt]{\textbf{Methods}}
            & \multicolumn{2}{c}{Sintel}
            & \multicolumn{2}{c}{KITTI} \\
            \cmidrule(lr){2-3} \cmidrule(lr){4-5}
            & Clean $\downarrow$ & Final $\downarrow$ & EPE $\downarrow$ & Fl-all $\downarrow$ \\
            \midrule
            PWC-Net \cite{pwc_net}     & 2.55 & 3.93 & 10.4 & 33.7 \\
            RAFT \cite{raft}        & 1.43 & 2.71 & 5.04 & 17.4 \\
            GMA  \cite{gma}       & 1.30 & 2.74 & 4.69 & 17.1 \\
            SKFlow \cite{skflow}     & 1.22 & 2.46 & 4.27 & 15.5 \\
            FlowFormer \cite{flowformer}  & \second{1.01} & \second{2.40} & 4.09 & 14.7 \\
            DIP \cite{dip}        & 1.30 & 2.82 & 4.29 & 13.7 \\
            EMD-L \cite{emd_l}      & \best{0.88} & 2.55 & 4.12 & 13.5 \\
            CRAFT \cite{craft}      & 1.27 & 2.79 & 4.88 & 17.5 \\
            RPKNet \cite{rpknet}     & 1.12 & 2.45 & -    & 13.0 \\
            GMFlowNet \cite{gmflownet}  & 1.14 & 2.71 & 4.24 & 15.4 \\
            SEA-RAFT(M) \cite{sea_raft} & 1.21 & 4.04 & 4.29 & 14.2 \\
            SEA-RAFT(L) \cite{sea_raft} & 1.19 & 4.11 & \second{3.62} & \second{12.9} \\
            \midrule
            MatchAttention (Ours) & 1.10 & \best{2.16} & \best{3.42} & \best{9.62} \\
            \bottomrule[1pt]
        \end{tabular*}
\end{table}

\begin{table}[]
\revisionon
    \centering
    \caption{Zero-shot optical-flow generalization on the high-resolution Spring dataset (bad 1.0 metric). Training set baseline results are evaluated using PTFlow\protect\footnote{https://github.com/hmorimitsu/ptlflow}; test set results are from the official Spring benchmark~\cite{spring}. ``--'' denotes methods that did not submit zero-shot results to the Spring test set. $\dagger$ denotes out-of-memory at inference.}
    \label{tab:s2:zero_shot_flow_spring}
        \begin{tabular*}{0.95\columnwidth}{@{\extracolsep{\fill}}lccc}
            \toprule[1pt]
            \multirow{2}*[-2pt]{\textbf{Methods}}
            & \multicolumn{2}{c}{Spring train} & Spring test \\
            \cmidrule(lr){2-3} \cmidrule(lr){4-4}
            & 2K $\downarrow$ & 4K $\downarrow$ & 2K $\downarrow$ \\
            \midrule
            PWC-Net \cite{pwc_net}     & 4.94 & 9.78 & 82.2 \\
            RAFT \cite{raft}        & 3.85 & 10.7 & 6.79 \\
            GMA  \cite{gma}       & 3.75 & $\dagger$ & 7.07 \\
            SKFlow \cite{skflow}     & 3.79 & $\dagger$ & -- \\
            FlowFormer \cite{flowformer}  & 3.78 & 9.53 & 6.51 \\
            DIP \cite{dip}        & 3.64 & 9.53 & -- \\
            GMFlow+ \cite{xu2023unifying}     & 5.72 & $\dagger$ & -- \\
            RPKNet \cite{rpknet}     & 3.28 & \second{7.35} & 4.80 \\
            MemFlow \cite{memflow}  & \second{3.24} & $\dagger$ & \second{5.76} \\
            SEA-RAFT(M) \cite{sea_raft} & 3.47 & 8.22 & -- \\
            \midrule
            MatchAttention (Ours) & \best{3.04} & \best{7.08} & \best{4.58} \\
            \bottomrule[1pt]
        \end{tabular*}
\end{table}

\noindent\textbf{Sintel / KITTI 2015.} Table~\ref{tab:s1:zero_shot_flow} shows that MatchAttention remains competitive with strong iterative and global-attention-based baselines under the same zero-shot protocol, while being trained on Scene Flow rather than on the standard C+T curriculum.

\noindent\textbf{Spring (2K / 4K).} Table~\ref{tab:s2:zero_shot_flow_spring} reports the results on Spring under the standard bad 1.0 metric. MatchAttention obtains the best result on Spring-train-2K, Spring-train-4K, and the public test set (2K), while several compared baselines run out of memory at 4K.

\revisionon
\section{Inference Profiling and Complexity Analysis}
\label{sec:s:profiling}

The linear complexity of MatchAttention follows from the constant-size sampling window analyzed in Section~\refmain{sec:3.1:matchattention} of the main paper. This section empirically compares MatchAttention with global attention at the operator level and profiles MatchAttentionRT end-to-end latency across input resolutions on a Jetson Orin NX.

\begin{table}[]
\revisionon
    \centering
    \caption{Inference comparison between MatchAttention and global attention at 256 channels (FP32, linear layers excluded).}
    \label{tab:s3:attention_compare}
        \begin{tabular*}{0.95\columnwidth}{@{\extracolsep{\fill}}lccc}
            \toprule[1pt]
            \multirow{2}{*}{Methods} & Token & Memory & Latency \\
            & resolution & (MB) $\downarrow$ & (ms) $\downarrow$ \\
            \midrule
            Global attention & $196 \times 196$ & 17630 & 27.9 \\
            \midrule
            \multirow{5}{*}{MatchAttention} & $196 \times 196$ & 870 & 1.4 \\
            & $512 \times 512$ & 2448 & 4.2 \\
            & $768 \times 768$ & 4652 & 7.7 \\
            & $1024 \times 1024$ & 7804 & 12.5 \\
            & $2048 \times 2048$ & 29464 & 43.1 \\
            \bottomrule[1pt]
        \end{tabular*}
\end{table}

\noindent\textbf{Operator-level scaling.} Table~\ref{tab:s3:attention_compare} compares MatchAttention with global attention \cite{Vaswani2017attention} under identical conditions (FP32, linear layers excluded, single RTX~4090). At a token resolution of $196\!\times\!196$, MatchAttention reduces memory by a factor of $20.3\times$ and latency by $19.9\times$. The gap arises from the quadratic memory footprint of global attention, which grows with $\mathcal{O}(N^2)$ in the number of tokens $N$, against the $\mathcal{O}(N)$ scaling of MatchAttention. At $256\!\times\!256$ tokens, global attention exceeds the available device memory, whereas MatchAttention scales to $2048\!\times\!2048$ tokens and completes a forward pass in 43.1\,ms.

\begin{table}[!t]
\revisionon
    \centering
    \caption{Latency and throughput of MatchAttentionRT on Jetson Orin NX 16\,GB at multiple input resolutions (FP16, \texttt{torch.compile}).}
    \label{tab:s4:edge_scan}
    \begin{tabular*}{0.95\columnwidth}{@{\extracolsep{\fill}}lccc}
        \toprule[1pt]
        \textbf{Resolution} & \textbf{FLOPs} $\downarrow$ & \textbf{Latency} $\downarrow$ & \textbf{FPS} $\uparrow$ \\
        \midrule
        $1024 \times 512$  & 62.5\,G  & 79.1\,ms   & 12.6 \\
        $1536 \times 864$  & 160.3\,G         & 172.6\,ms         & 5.8 \\
        $1920 \times 1088$ & 255.5\,G         & 280.1\,ms         & 3.6 \\
        \bottomrule[1pt]
    \end{tabular*}
\end{table}

\noindent\textbf{Edge deployment.} Table~\ref{tab:s4:edge_scan} reports the latency of MatchAttentionRT on the Jetson Orin NX~16\,GB at $1024\times512$, $1536\times864$, and $1920\times1088$ (FP16, \texttt{torch.compile}). From the lowest to the highest tested resolution, latency increases approximately linearly with pixel count, consistent with the $\mathcal{O}(N)$ operator complexity. No per-device tuning or quantization is applied beyond the three hardware-aware optimizations described in Section~\refmain{sec:3.4:acceleration}.

\revisionon
\section{Qualitative Comparison of Real-Time Zero-Shot Stereo}
\label{sec:s:realtime_visual}

Section~\refmain{sec:4.4:zero_shot} and Table~\refmain{tab:5:realtime_zero_shot} report the quantitative zero-shot comparison of real-time stereo methods. Here we provide a qualitative comparison on high-resolution Middlebury-F scenes to show the error patterns behind those metrics. We compare MatchAttentionRT with two real-time methods that support an adjustable disparity search range (Fast-FSD-416~\cite{fastfoundationstereo} and RT-Monster++-416~\cite{monster++}, both with default settings) and one compact feed-forward method with a fixed disparity search range (LiteAnyStereo~\cite{liteanyStereo}).

\begin{figure*}[!t]
\revisionon
    \centering
    \subfloat[]{\includegraphics[width=0.195\textwidth]{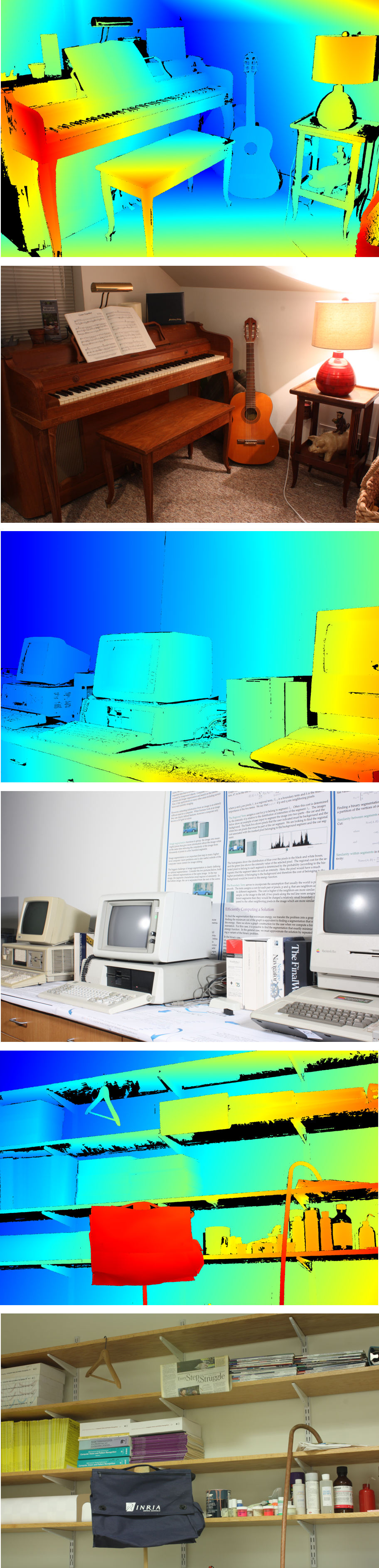}}
    \hfill
    \subfloat[]{\includegraphics[width=0.195\textwidth]{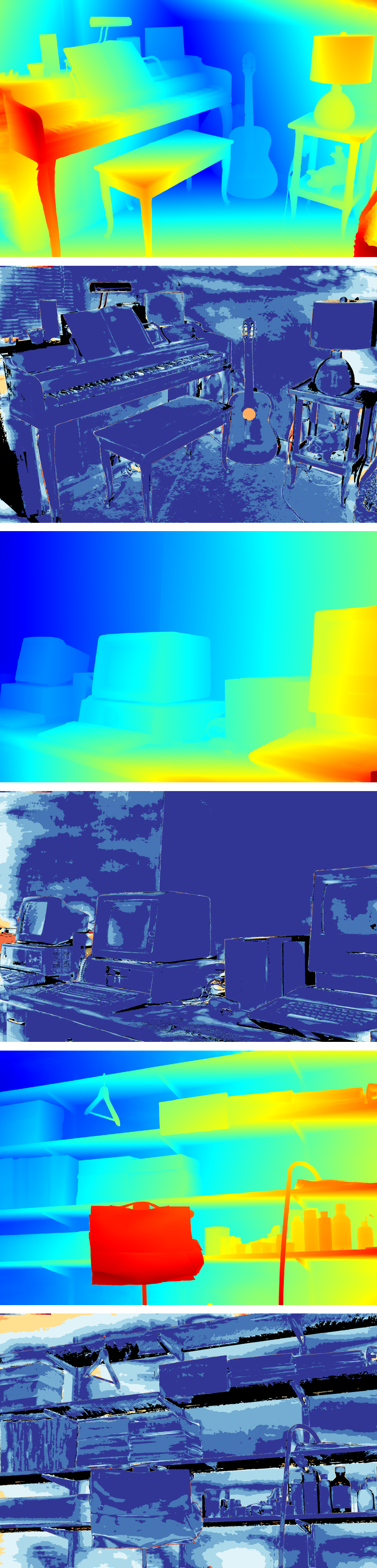}}
    \hfill
    \subfloat[]{\includegraphics[width=0.195\textwidth]{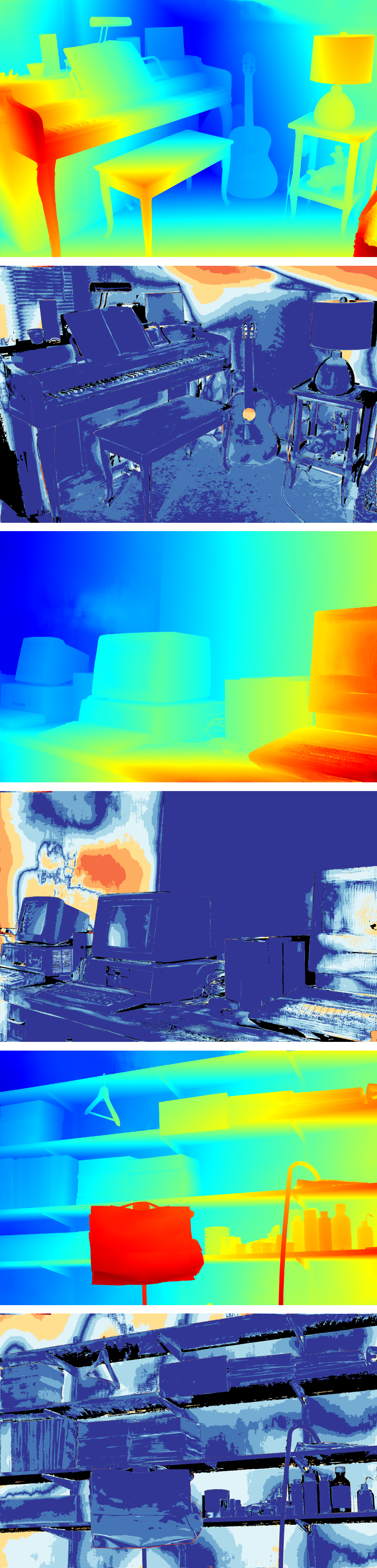}}
    \hfill
    \subfloat[]{\includegraphics[width=0.195\textwidth]{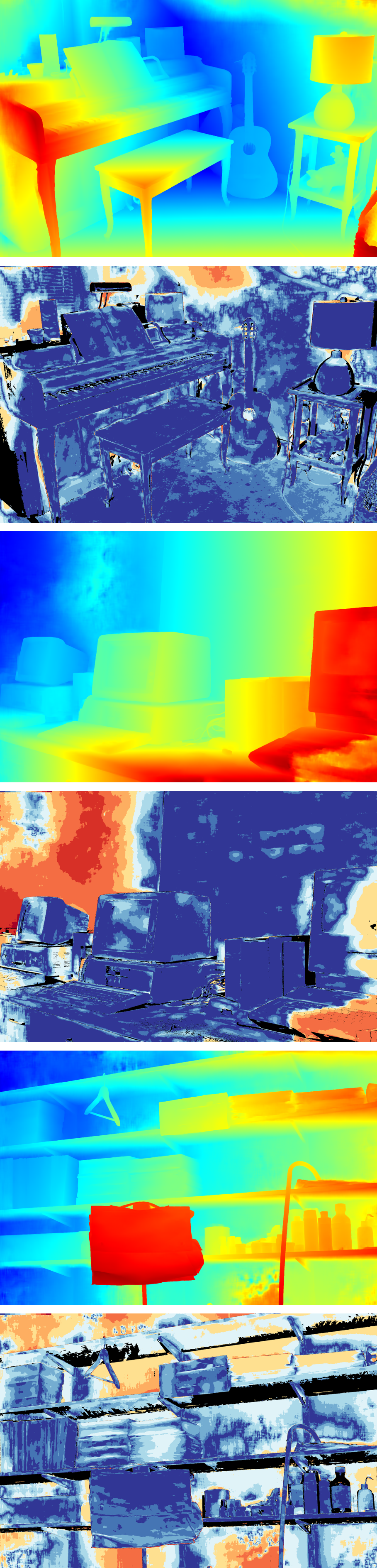}}
    \hfill
    \subfloat[]{\includegraphics[width=0.195\textwidth]{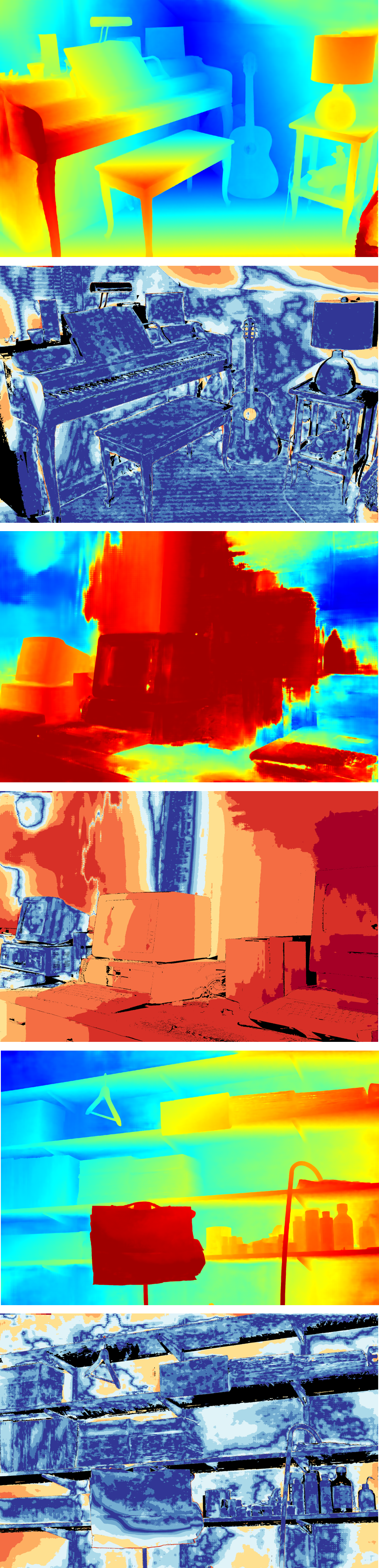}}
    \caption{Comparison on Piano, Vintage, and Shelves image pairs of the Middlebury-F benchmark. (a) Color reference images and ground truth disparity maps. (b)--(e) Disparity and error maps of MatchAttentionRT (EPE: 0.61, 0.73, 1.01), Fast-FSD-416~\cite{fastfoundationstereo} (EPE: 1.18, 2.73, 1.18), RT-Monster++-416~\cite{monster++} (EPE: 1.09, 8.78, 3.30), and LiteAnyStereo~\cite{liteanyStereo} (EPE: 1.63, 111.87, 2.20), respectively.}
    \label{fig:s:realtime_visual}
\end{figure*}

As shown in Fig.~\ref{fig:s:realtime_visual}, MatchAttentionRT produces cleaner disparity maps and lower EPE around the lamp highlight in Piano, where strong specular response makes correspondence difficult for the compared methods. On Vintage, it also produces a more accurate and spatially stable disparity over the large, weakly textured white wall. On Shelves, it again achieves the lowest EPE on the largely textureless shelf surfaces. LiteAnyStereo~\cite{liteanyStereo} exhibits large errors on the Vintage wall because local disparities exceed its fixed maximum disparity of 192 pixels at inference, consistent with Section~\refmain{sec:4.4:zero_shot}. Fast-FSD-416~\cite{fastfoundationstereo} and RT-Monster++-416~\cite{monster++} allow the disparity search range to be adjusted at inference and are therefore not limited by this fixed range, but their errors on Vintage remain substantially higher than those of MatchAttentionRT. On Shelves, the maximum ground-truth disparity is approximately 202 pixels, close to the maximum disparity of LiteAnyStereo, yet MatchAttentionRT still obtains the lowest EPE among all compared methods. These observations are consistent with Table~\refmain{tab:5:realtime_zero_shot} and indicate that the accuracy gap is driven by the explicit matching constraint of MatchAttention rather than by disparity search range alone.

\revisionon
\section{Full Middlebury Leaderboard}
\label{sec:s:benchmarks}

The main paper (Section~\refmain{sec:4.3:benchmark}, Table~\refmain{tab:2:middv3_perscene}) reports Middlebury V3 per-scene results on bad~0.5 and bad~1.0 only. Table~\ref{tab:s5:middv3_perscene_full} below extends this comparison to all six error metrics (bad~0.5, bad~1.0, bad~2.0, bad~4.0, EPE, and RMS) and the complete per-scene leaderboard.

% Auto-generated by scripts/build_middv3_anchor.py — do not edit by hand.
\begin{table*}[!p]
\revisionon
    \centering
    \caption{Per-scene accuracy on the Middlebury~V3 benchmark across all six error metrics (Noc mask, full resolution). The same methods as Tab.~\refmain{tab:2:middv3_perscene} of the main paper are listed. Within each panel, baselines are ordered by weighted-average ascending; MatchAttentionXL is listed in the last row. Best per column in yellow bold, second-best in soft blue.}
    \label{tab:s5:middv3_perscene_full}
        \scriptsize
        \setlength{\tabcolsep}{1.4pt}
        \begin{tabular*}{\textwidth}{@{\extracolsep{\fill}}lcccccccccccccccc}
            \toprule[1pt]
            \multirow{2}{*}{\textbf{Method}} & \multirow{2}{*}{\textbf{Avg}} & \multicolumn{15}{c}{\textbf{Per-scene} $\downarrow$} \\
            \cmidrule(lr){3-17}
            & & Aust & AustP & Bcyc & Clas & ClaE & Comp & Crus & CruP & Djmb & DjmL & Hoop & Lvgr & Nkub & Plnt & Stair \\
            \midrule
            \multicolumn{17}{l}{\textit{(a) Bad 0.5\,(\%) $\downarrow$}} \\
            \midrule
            S2M2 \cite{s2m2} & \second{22.1} & 23.4 & 17.2 & \best{8.28} & 18.6 & \best{45.0} & 45.2 & 27.0 & 26.1 & 13.7 & 19.4 & 20.1 & 13.0 & \second{30.0} & \best{21.4} & \best{3.92} \\
            FoundationStereo \cite{foundationstereo} & 22.5 & \best{11.1} & 10.7 & 8.66 & 17.9 & 61.5 & \best{36.2} & \second{21.3} & \second{20.7} & \second{7.82} & \second{16.9} & \second{18.2} & 12.1 & 50.7 & 36.6 & \second{10.5} \\
            MGS-Stereo \cite{mgsstereo} & 23.4 & 17.0 & 15.7 & 9.43 & 16.4 & \second{52.1} & 38.4 & 32.6 & 32.7 & 12.4 & 24.0 & 24.5 & 13.2 & \best{29.6} & 26.8 & 11.8 \\
            AIO-Stereo \cite{AIO} & 24.0 & 16.6 & 14.7 & 10.6 & \second{13.0} & 59.8 & 38.5 & 36.5 & 35.2 & 7.92 & 24.8 & 22.4 & 13.7 & 32.2 & 29.4 & 13.6 \\
            Selective-IGEV \cite{selective} & 24.6 & 15.2 & 14.6 & 10.3 & 14.2 & 61.5 & 38.7 & 36.8 & 35.6 & 9.93 & 29.0 & 25.3 & 12.2 & 35.6 & \second{25.6} & 17.6 \\
            DEFOM-Stereo \cite{defom_stereo} & 25.2 & \second{11.6} & 21.4 & 9.65 & 17.8 & 55.0 & 37.3 & 34.6 & 34.1 & 12.1 & 24.6 & 27.0 & \second{11.6} & 32.9 & 38.0 & 13.7 \\
            StereoAnywhere \cite{stereo_anywhere} & 27.3 & 27.2 & \second{7.88} & 12.1 & 17.4 & 58.4 & 45.0 & 37.0 & 32.3 & 8.72 & 28.4 & 30.0 & 13.4 & 50.0 & 35.3 & 20.3 \\
            MatchAttentionXL & \best{21.3} & 15.8 & \best{7.79} & \second{8.61} & \best{10.1} & 55.3 & \second{37.2} & \best{16.5} & \best{18.0} & \best{5.89} & \best{9.76} & \best{13.9} & \best{8.17} & 48.9 & 47.7 & 19.3 \\
            \midrule
            \multicolumn{17}{l}{\textit{(b) Bad 1.0\,(\%) $\downarrow$}} \\
            \midrule
            S2M2 \cite{s2m2} & \second{3.57} & \best{2.81} & \best{2.33} & 3.50 & 2.33 & \best{3.18} & 2.07 & \second{2.58} & \second{2.75} & 1.97 & \second{4.12} & 9.24 & 4.44 & 9.36 & \best{3.18} & \best{0.87} \\
            FoundationStereo \cite{foundationstereo} & 4.39 & 4.23 & 2.74 & \second{3.13} & \second{2.10} & 11.4 & \best{1.57} & 2.83 & 2.86 & 1.84 & 6.34 & \second{6.97} & 5.30 & 11.7 & 4.75 & \second{3.13} \\
            MGS-Stereo \cite{mgsstereo} & 5.69 & 4.89 & 3.14 & 4.85 & 3.28 & 16.3 & 3.54 & 5.83 & 5.72 & \second{1.83} & 6.04 & 11.1 & 4.26 & 12.1 & 5.27 & 4.17 \\
            DEFOM-Stereo \cite{defom_stereo} & 5.81 & 4.57 & 3.92 & 4.79 & 3.35 & 14.3 & 3.04 & 5.67 & 5.50 & 2.41 & 7.54 & 15.3 & \second{3.98} & 11.3 & 5.71 & 4.22 \\
            AIO-Stereo \cite{AIO} & 6.08 & 4.54 & 3.16 & 5.98 & 2.76 & 16.2 & 3.52 & 7.97 & 7.67 & 2.29 & 10.4 & 10.7 & 4.64 & \best{8.39} & 5.34 & 6.76 \\
            Selective-IGEV \cite{selective} & 6.53 & 4.68 & 3.16 & 5.32 & 3.86 & 17.0 & 3.60 & 8.84 & 8.12 & 2.74 & 11.5 & 10.7 & 5.12 & \second{9.05} & 5.86 & 8.00 \\
            StereoAnywhere \cite{stereo_anywhere} & 7.99 & 13.0 & 3.49 & 5.40 & 7.96 & 28.2 & 3.51 & 8.70 & 6.20 & 2.37 & 12.1 & 13.1 & 5.83 & 15.0 & 5.60 & 5.16 \\
            MatchAttentionXL & \best{3.49} & \second{3.20} & \second{2.47} & \best{2.43} & \best{1.61} & \second{6.20} & \second{1.91} & \best{2.43} & \best{2.21} & \best{1.33} & \best{2.60} & \best{4.58} & \best{1.87} & 12.3 & \second{4.70} & 4.26 \\
            \midrule
            \multicolumn{17}{l}{\textit{(c) Bad 2.0\,(\%) $\downarrow$}} \\
            \midrule
            S2M2 \cite{s2m2} & \best{1.15} & \best{1.29} & \best{1.23} & \second{1.27} & \best{0.40} & \best{0.45} & 0.59 & \best{0.67} & \best{0.62} & \best{0.45} & \second{1.28} & \second{2.80} & \second{1.37} & \second{3.60} & \best{1.12} & \best{0.25} \\
            FoundationStereo \cite{foundationstereo} & 1.84 & 2.46 & 1.71 & 1.36 & 0.79 & 5.19 & \best{0.53} & \second{0.93} & 0.84 & 0.93 & 2.41 & 3.39 & 3.45 & \best{3.28} & 1.82 & \second{1.17} \\
            MGS-Stereo \cite{mgsstereo} & 2.16 & 2.62 & \second{1.58} & 2.20 & 0.76 & 6.45 & 1.04 & 1.39 & 1.08 & \second{0.67} & 1.73 & 4.32 & 1.43 & 6.29 & 1.83 & 2.30 \\
            AIO-Stereo \cite{AIO} & 2.36 & 2.38 & 1.71 & 3.22 & 0.85 & 5.83 & 1.24 & 1.42 & 1.32 & 1.03 & 4.49 & 4.81 & 2.43 & 3.61 & 2.12 & 3.63 \\
            DEFOM-Stereo \cite{defom_stereo} & 2.39 & 2.82 & 2.21 & 1.53 & 1.01 & 5.24 & 0.88 & 1.40 & 1.14 & 0.85 & 2.64 & 9.10 & 2.18 & 5.50 & 2.49 & 1.67 \\
            Selective-IGEV \cite{selective} & 2.51 & 2.54 & 1.86 & 2.51 & 1.12 & 7.22 & 1.23 & 1.36 & 1.17 & 1.16 & 4.48 & 4.83 & 2.99 & 3.79 & 2.26 & 4.72 \\
            StereoAnywhere \cite{stereo_anywhere} & 3.69 & 7.34 & 2.23 & 2.23 & 5.12 & 18.1 & 0.90 & 2.16 & 1.43 & 1.25 & 5.73 & 4.95 & 2.66 & 6.89 & 2.28 & 1.86 \\
            MatchAttentionXL & \second{1.29} & \second{1.99} & 1.73 & \best{0.98} & \second{0.48} & \second{0.94} & \second{0.55} & 0.93 & \second{0.77} & 0.67 & \best{1.22} & \best{1.31} & \best{0.98} & 4.16 & \second{1.39} & 1.45 \\
            \midrule
            \multicolumn{17}{l}{\textit{(d) Bad 4.0\,(\%) $\downarrow$}} \\
            \midrule
            S2M2 \cite{s2m2} & \best{0.54} & \best{1.00} & \best{0.98} & \best{0.74} & \best{0.30} & \best{0.29} & \second{0.32} & \best{0.42} & \best{0.41} & \best{0.25} & \second{0.62} & \second{1.06} & \best{0.65} & \best{0.34} & \best{0.75} & \best{0.16} \\
            MGS-Stereo \cite{mgsstereo} & 0.97 & \second{1.51} & \second{1.18} & \second{0.77} & 0.39 & 2.95 & 0.40 & 0.70 & 0.56 & 0.45 & 0.90 & 1.78 & 0.95 & 1.31 & 1.08 & 1.65 \\
            FoundationStereo \cite{foundationstereo} & 1.04 & 1.58 & 1.36 & 1.05 & 0.44 & 3.52 & 0.36 & \second{0.59} & \second{0.51} & 0.62 & 1.00 & 1.93 & 1.91 & 0.85 & 1.02 & \second{0.59} \\
            DEFOM-Stereo \cite{defom_stereo} & 1.22 & 1.93 & 1.60 & 1.04 & 0.51 & 2.84 & 0.47 & 0.81 & 0.62 & 0.48 & 1.05 & 3.74 & 1.24 & 1.71 & 1.67 & 0.74 \\
            AIO-Stereo \cite{AIO} & 1.26 & 1.60 & 1.26 & 0.85 & 0.50 & 3.85 & 0.62 & 0.80 & 0.63 & 0.55 & 1.69 & 3.23 & 1.74 & 1.18 & 1.35 & 2.14 \\
            Selective-IGEV \cite{selective} & 1.36 & 1.76 & 1.40 & 0.88 & 0.51 & 4.12 & 0.59 & 0.78 & 0.61 & 0.68 & 1.76 & 2.78 & 1.99 & 1.24 & 1.37 & 3.58 \\
            StereoAnywhere \cite{stereo_anywhere} & 2.17 & 4.47 & 1.68 & 0.93 & 4.01 & 13.8 & 0.47 & 0.90 & 0.65 & 0.94 & 2.12 & 2.33 & 1.24 & 3.11 & 1.36 & 0.90 \\
            MatchAttentionXL & \second{0.67} & 1.58 & 1.38 & 0.82 & \second{0.34} & \second{0.39} & \best{0.31} & 0.66 & 0.52 & \second{0.38} & \best{0.61} & \best{0.65} & \second{0.69} & \second{0.46} & \second{0.87} & 0.70 \\
            \midrule
            \multicolumn{17}{l}{\textit{(e) EPE (px) $\downarrow$}} \\
            \midrule
            S2M2 \cite{s2m2} & \second{0.69} & 1.30 & 1.23 & 0.60 & \second{0.53} & \best{0.66} & 0.58 & \second{0.67} & \second{0.65} & 0.33 & \second{0.42} & \second{0.72} & \second{0.47} & \second{0.99} & \best{0.82} & \best{0.30} \\
            MGS-Stereo \cite{mgsstereo} & 0.74 & \second{1.16} & \second{1.11} & \best{0.52} & 0.55 & 1.07 & 0.52 & 0.91 & 0.71 & 0.39 & 0.49 & 0.93 & 0.53 & 1.05 & \second{0.92} & 0.53 \\
            FoundationStereo \cite{foundationstereo} & 0.78 & 1.45 & 1.37 & 0.69 & 0.55 & 1.14 & \second{0.48} & \best{0.66} & \best{0.63} & 0.31 & 0.42 & 0.83 & 0.63 & 1.24 & 1.00 & \second{0.52} \\
            DEFOM-Stereo \cite{defom_stereo} & 0.79 & \best{1.14} & 1.16 & \second{0.53} & 0.57 & 1.07 & 0.50 & 0.76 & 0.73 & 0.39 & 0.53 & 1.73 & 0.53 & 1.12 & 1.06 & 0.55 \\
            AIO-Stereo \cite{AIO} & 0.85 & 1.37 & 1.26 & 0.60 & 0.58 & 1.23 & 0.54 & 1.05 & 0.76 & \second{0.30} & 0.58 & 1.41 & 0.70 & 1.07 & 1.12 & 0.66 \\
            Selective-IGEV \cite{selective} & 0.91 & 1.44 & 1.40 & 0.60 & 0.69 & 1.36 & 0.55 & 1.06 & 0.76 & 0.32 & 0.63 & 1.24 & 0.70 & 1.09 & 1.18 & 1.46 \\
            StereoAnywhere \cite{stereo_anywhere} & 0.93 & 1.50 & \best{1.06} & 0.60 & 0.96 & 2.77 & 0.59 & 0.82 & 0.75 & 0.39 & 0.70 & 1.00 & 0.53 & 1.71 & 0.98 & 0.59 \\
            MatchAttentionXL & \best{0.68} & 1.38 & 1.20 & 0.57 & \best{0.49} & \second{0.82} & \best{0.47} & 0.68 & 0.67 & \best{0.29} & \best{0.34} & \best{0.47} & \best{0.40} & \best{0.92} & 0.94 & 0.65 \\
            \midrule
            \multicolumn{17}{l}{\textit{(f) RMS (px) $\downarrow$}} \\
            \midrule
            DEFOM-Stereo \cite{defom_stereo} & \second{5.81} & \best{9.51} & \second{9.43} & \best{4.32} & \second{4.60} & \second{5.08} & \second{2.11} & \best{5.83} & \best{5.74} & 1.75 & 1.85 & 10.9 & 3.56 & \second{11.2} & \second{7.80} & 5.25 \\
            MGS-Stereo \cite{mgsstereo} & 5.91 & 9.95 & 9.65 & \second{4.44} & \best{4.56} & 5.29 & 2.27 & 8.27 & \second{5.93} & 1.54 & 1.65 & 6.58 & 3.94 & 11.3 & 7.90 & \best{4.60} \\
            S2M2 \cite{s2m2} & 6.01 & 10.9 & 10.9 & 5.08 & 4.71 & \best{4.78} & 2.51 & \second{6.11} & 6.08 & \best{1.11} & \best{1.36} & 6.58 & \second{3.15} & 13.5 & 7.94 & \second{4.65} \\
            StereoAnywhere \cite{stereo_anywhere} & 6.18 & \second{9.51} & \best{9.27} & 4.66 & 5.30 & 7.47 & 2.20 & 6.25 & 6.19 & 1.99 & 2.40 & 7.13 & 3.53 & 14.2 & 7.95 & 4.90 \\
            FoundationStereo \cite{foundationstereo} & 6.48 & 12.6 & 12.4 & 5.80 & 4.94 & 5.51 & 2.45 & 6.28 & 6.17 & 1.24 & \second{1.43} & \second{6.40} & 4.09 & 13.7 & 8.56 & 4.82 \\
            AIO-Stereo \cite{AIO} & 6.78 & 11.5 & 11.1 & 4.76 & 5.37 & 5.59 & 2.48 & 10.5 & 6.12 & 1.32 & 1.85 & 9.01 & 4.46 & 12.6 & 8.85 & 6.28 \\
            Selective-IGEV \cite{selective} & 7.26 & 12.2 & 12.2 & 4.82 & 6.86 & 7.25 & 2.50 & 10.6 & 6.04 & 1.44 & 2.09 & 8.10 & 4.54 & 12.6 & 9.36 & 9.79 \\
            MatchAttentionXL & \best{5.66} & 10.9 & 10.3 & 4.78 & 5.31 & 5.54 & \best{2.06} & 6.71 & 6.65 & \second{1.11} & 1.54 & \best{4.56} & \best{3.00} & \best{9.08} & \best{7.47} & 5.91 \\
            \bottomrule[1pt]
        \end{tabular*}
\end{table*}

Across the six metrics in Table~\ref{tab:s5:middv3_perscene_full}, MatchAttentionXL is the only method that ranks in the top two on every metric. It is the best on bad~0.5, bad~1.0, EPE, and RMS, and second on bad~2.0 and bad~4.0. The per-scene breakdown confirms that the advantage is consistent rather than driven by a few scenes, as MatchAttentionXL obtains the per-scene best on a majority of the 15 scenes under bad~0.5, bad~1.0, and EPE.

\end{document}